\documentclass[10pt,twocolumn,letterpaper]{article} 

\usepackage{avss}
\usepackage{times}
\usepackage{epsfig}
\usepackage{graphicx}
\usepackage{amsmath}
\usepackage{multirow}
\usepackage{enumitem}
\usepackage{float}
\usepackage{booktabs}

\usepackage[justification=centering]{caption}
\usepackage{amssymb}
\usepackage{xcolor}

\usepackage[pagebackref=false,breaklinks=true,letterpaper=true,colorlinks,bookmarks=false]{hyperref}

\usepackage{multibib}
\newcites{supp}{Supplementary References}

\avssfinalcopy 

\ifavssfinal\fi
\begin{document}

\title{
A Data-Centric Approach to Pedestrian Attribute Recognition:\\
Synthetic Augmentation via Prompt-driven Diffusion Models
}
\author{Alejandro Alonso$^{1,2,3}$, Sawaiz A. Chaudhry$^{1,2,3}$, Juan C. SanMiguel$^{1,}$, \\ \'{A}lvaro Garc\'{i}a-Mart\'{i}n$^{1}$, Pablo Ayuso-Albizu$^{1}$, Pablo Carballeira$^{1}$\\
$^{1}$Autonomous University of Madrid, Madrid, Spain. 
$^{2}$University of Bordeaux, Bordeaux, France.\\
$^{3}$P\'{a}zm\'{a}ny P\'{e}ter Catholic University, Budapest, Hungary.\\
\tt\small \{alejandro.alonsog, sawaiz.chaudhry\}@estudiante.uam.es, \\ 
\tt\small\{juancarlos.sanmiguel, alvaro.garcia, pablo.ayuso, pablo.carballeira\}@uam.es
}

\maketitle

\begin{abstract}
   Pedestrian Attribute Recognition (PAR) is a challenging task as models are required to generalize across numerous attributes in real-world data. Traditional approaches focus on  complex methods, yet recognition performance is often constrained by training dataset limitations, particularly the under-representation of certain attributes. In this paper, we propose a data-centric approach to improve PAR by synthetic data augmentation guided by textual descriptions. First, we define a protocol to  identify weakly recognized attributes across multiple datasets. Second, we propose a prompt-driven pipeline that leverages diffusion models to generate synthetic pedestrian images while preserving the consistency of PAR datasets. Finally, we derive a strategy to seamlessly incorporate synthetic samples into training data, which considers prompt-based annotation rules and modifies the loss function. Results on popular PAR datasets demonstrate that our approach not only boosts recognition of underrepresented attributes but also improves overall model performance beyond the targeted attributes. Notably, this approach strengthens zero-shot generalization without requiring architectural changes of the model, presenting an efficient and scalable solution to improve the recognition of attributes of pedestrians in the real world.
\end{abstract}

\section{Introduction}
Pedestrian Attribute Recognition (PAR) aims to identify multiple soft biometric attributes of individuals in human-centric AI systems. Despite significant advances in deep learning-based approaches, achieving stable and consistent performance across all attributes remains a challenge \cite{jin2024pedestrian}. Most state-of-the-art models struggle with imbalanced and under-represented attribute distributions derived from real-world data \cite{zhou2025heterogeneous}, leading to suboptimal recognition for certain attributes, even with sophisticated learning architectures. Many attributes appear infrequently or with limited variability, making it difficult for models to learn generalizable representations. Additionally, in zero-shot learning scenarios, where models must recognize attributes of people identities that were not seen during training, further expose the limitations of conventional PAR approaches. 

Recently, generative AI has emerged as a solution to address data scarcity in computer vision tasks \cite{chen2024comprehensive}. Methods based on Generative Adversarial Networks, Variational Autoencoders, and Diffusion Models (DMs) have been extensively explored for synthesizing high-quality images to support data augmentation and zero-shot learning. Among these, DMs have recently gained prominence to generate highly realistic and diverse images while providing fine-grained control over content \cite{croitoru2023diffusion}. Despite the success of DMs, their potential for augmenting PAR datasets remains relatively unexplored, presenting a promising avenue for improving attribute recognition performance.

In this paper, we propose a data-centric approach to improve PAR performance through synthetic data. Instead of increasing model complexity, we enhance data representation quality by generating synthetic pedestrian images to address the imbalance of under-represented attributes. Extensive experiments on popular datasets (RAPv1, RAPv2 \cite{li2018richly}, and RAPzs \cite{jia2021rethinking}) show that our method boosts recognition of augmented attributes while preserving overall robustness.

Our contributions are threefold: (1) \textbf{A systematic protocol} to identify weakly recognized attributes across multiple datasets; (2) \textbf{A prompt-driven pipeline} leveraging diffusion models to generate synthetic pedestrian images while preserving dataset consistency; and (3) \textbf{An integration strategy for PAR} data augmentation that modifies the loss function to seamlessly incorporate synthetic samples into training, ensuring smooth adaptation of new data.  


\section{Related Work}

\subsection{PAR: Methods and Datasets}
Recent approaches for Pedestrian Attribute Recognition (PAR) integrate attribute-specific characteristics beyond architectural improvements \cite{wang2022pedestrian}, where ResNet is a popular backbone \cite{ jia2021rethinking, kirillova2022visual, galiyawala2021dsa}. For example, discrimination can be enhanced by introducing location-aware associations of attributes with body regions \cite{SHEN2024110194}. Similarly, attention mechanisms \cite{WENG2023140} can be used to model attribute correlations, enabling to refine predictions. Dataset biases can also introduce inconsistencies for learning these attribute relationships, being addressed by information-theoretic regularization \cite{zhou2023solution}. Multi-task transformers have been also proposed to capture global and local relations \cite{fan2023parformer}. 
In summary, many recent approaches focus on algorithmic complexity or inter-attribute correlations to enhance PAR performance.

These complex PAR approaches rely on datasets with challenges for data distribution and generalization. For example, PETA \cite{peta_dataset} and PA100K \cite{pa100k_dataset} cover both indoor and outdoor environments, while RAPv1 
and its extension RAPv2 \cite{li2018richly} focus on indoor surveillance with a richer set of attributes. While efforts such as PETAZs and RAPZs \cite{jia2021rethinking} have aimed to address zero-shot learning, PAR datasets remain highly imbalanced, with certain attributes severely underrepresented. Consequently, models trained on them struggle to generalize, and leveraging inter-attribute relationships proves ineffective, leading to biased learning.

\subsection{Text-guided Synthetic Image Generation}
Generative AI has become a powerful tool to mitigate dataset biases and augment training visual data \cite{lin2024robust}, with Generative Adversarial Networks (GANs) 
widely used to enhance data diversity. Early attempts to generate synthetic PAR data relied on unconditional GANs \cite{fabbri2017generative}, lacking fine-grained control over individual attributes. More recently, Diffusion Models (DMs) have surpassed GANs \cite{croitoru2023diffusion}, achieving high realism and diversity while allowing conditioning mechanisms, such as prompts or images. This flexibility has led to the rise of highly popular models, including DALL·E \cite{ramesh2021zero} and Stable Diffusion \cite{rombach2022high}. Hence, DMs arise as an option for generating PAR-like images with precise control over attributes, ensuring dataset consistency while adapting to diverse real-world scenarios.

Generating visual data through textual prompts is a powerful yet non-trivial task, especially when aiming to match the appearance of a target domain. DMs allow \textit{text-to-image generation} (txt2img), but standard prompting can lead to inconsistent attributes, unrealistic data, or domain mismatches \cite{schulhoff2024prompt}. Therefore, \textit{Prompt Engineering} refines prompt design to better align generated images with the target domain without modifying the model. For example, \textit{Wildcards and Template-Based Prompts} introduce controlled attribute variation ensuring dataset realism \cite{mo2024dynamic}, while \textit{Textual Inversion} learns custom token embeddings to generate domain-consistent data without retraining  \cite{gal2022image}. Conversely, \textit{Model Adaptation} enhances control in txt2img diffusion. Prompt-to-Prompt  \cite{hertz2022prompt} refines attributes via attention map adjustments, while LoRA fine-tuning \cite{hu2022lora} efficiently adapts models by updating a small parameter subset.

Despite PAR's challenges with class imbalance and domain bias, these techniques remain unexplored. This paper first examines prompt engineering to ensure realistic, attribute-consistent images before model adaptation.

\section{Proposed Approach}
Our proposal considers three steps to enhance Pedestrian Attribute Recognition (PAR) with synthetic images. First, we \textit{identify weakly recognized attributes} by analyzing class imbalance and recognition performance. Next, we generate \textit{synthetic pedestrian images} using a diffusion-based pipeline, ensuring realistic attribute representations. Finally, we \textit{integrate the synthetic data} into training by labeling the data and adapting the loss function. The next subsections outline the proposal; further details of the proposal are available in the supplementary material\footnote{\url{http://www-vpu.eps.uam.es/AVSS2025PAR}}.

\subsection{Identifying Weakly Recognized Attributes}\label{sec:weak-attributes}
We start by identifying the worst-performing attributes to apply our data-centric synthetic augmentation. Given one or more PAR training/test sets $D^j = \{(x^j_i, y^j_i), i = 1, 2, \ldots, N_j\}$, 
where $j$ indexes each dataset that contains $N_j$ pedestrian images. Each image is annotated with $M$ shared binary attributes $y^j_i \in \{0, 1\}^M$. Our goal is to determine which attributes require augmentation.

\vspace{-2mm}
\paragraph{Identification Criteria:} We identify weakly recognized attributes using three criteria: (1) class imbalance, (2) test-time performance, and (3) performance drop from training to testing. Attributes with few training samples are direct candidates for augmentation, while those with high test performance are deprioritized. Additionally, attributes showing a large drop in performance between training and testing indicate possible overfitting.

\vspace{-2mm}
\paragraph{Criteria Scores:} We employ a qualitative scoring system, assigning each attribute a score of 0 (worst case), 1 (medium), or 2 (best case). We propose a thresholding approach to define these categories, with threshold values derived from experimental analysis and insights from related work (e.g., attributes with fewer than 3\% of training samples are considered underrepresented, following \cite{jia2021rethinking}). Table \ref{table:attribute_aspects} summarizes the thresholds applied in this study. The final scores is obtained by adding the individual scores.

\begin{table}[t]
\centering
\caption{Thresholds for qualitative scoring of PAR attributes over one or multiple datasets.}
\vspace{-2mm}
\resizebox{\columnwidth}{!}{
\centering
\begin{tabular}{|c|c|c|c|}
\hline
\textbf{Categorization} & \textbf{Low training} & \textbf{Test} & \textbf{Drop in} \\
 \textbf{(Score)}& \textbf{images} & \textbf{performance} & \textbf{performance} \\ \hline
\textbf{Best (2)}                    & False ($>$3\%)                       & High ($>$80)                      & Small ($<$15)                       \\ \hline
\textbf{Medium (1)}                  & -                                    & Medium (50-80)                    & Medium (15-30)                      \\ \hline
\textbf{Worst (0)}                     & True ($<$3\%)                        & Low ($<$50)                       & Big ($>$30)                         \\ \hline
\end{tabular}
}
\label{table:attribute_aspects}
\end{table}

\begin{figure*}[h!]
\centering
\includegraphics[width=0.9\textwidth]{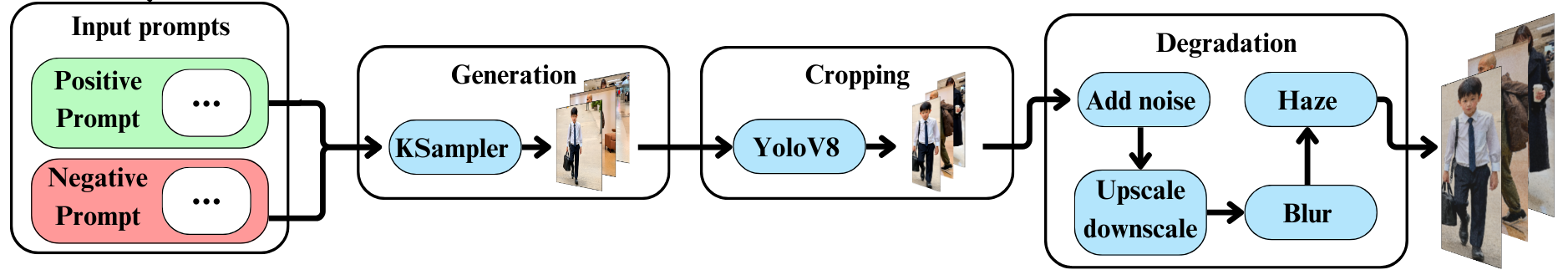}
\vspace{-2mm}
\caption{Proposed diffusion-based pipeline for synthetic PAR-like image generation.}
\label{fig:generation_diagram}
\vspace{-2mm}
\end{figure*}

\noindent\fbox{%
    \parbox{\linewidth}{%
        A highly detailed, ultra high definition image of a single pedestrian \textbf{\_\_poses\_\_} on \textbf{\_\_backgrounds\_\_}, with the pedestrian fully visible and centered in the frame. The pedestrian is visible from head to toe and is the primary focus, and the scene features realistic, natural colors. We have \textbf{\_\_views\_\_}. The subject is wearing a \textbf{\_\_styles\_\_}. \textbf{\_\_colors\_\_}. It can also be seen how the pedestrian \textbf{\_\_attributes\_\_}.
    }%
}
\vspace{-2mm}
\captionof{figure}{Positive prompt for PAR-like image generation.}
\label{fig:main_promt}
\setlength{\parindent}{1pc} 

\subsection{Synthetic PAR-like Image Generation} 
\label{sec:generation-pipeline}
After identifying the attributes to improve, we propose a pipeline based on text-to-image generation to ensure controlled outputs (see Figure \ref{fig:generation_diagram}), which is described as follows.

\vspace{-2mm}
\paragraph{Input Prompts for Attributes:} 
Designing effective prompts is crucial to generate diverse yet representative PAR dataset images. Our approach follows the \textit{Wildcards and Template-Based Prompts} strategy, where wildcards are dynamically replaced with pre-configured phrases based on predefined probabilities. This allows for richer, more complex prompts that better reflect dataset characteristics. We analyzed PAR datasets to define each wildcard, resulting in the positive prompt shown in Figure \ref{fig:main_promt}. Additionally, negative prompts were used to exclude undesired concepts. Thus, the resulting prompts provide a richer variability and complexity (see examples in Figure \ref{fig:generated_samples}).

\begin{figure}[t]
    \centering
    \includegraphics[width=0.95\columnwidth]{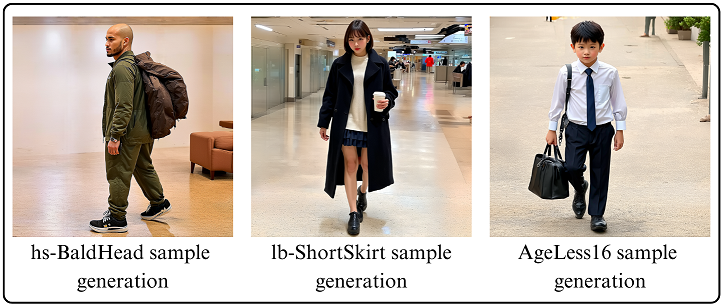}
    \caption{Samples generated with the proposed prompt for each one of the pedestrian attributes.}
    \label{fig:generated_samples}
\end{figure}

\vspace{-2mm}
\paragraph{Image Generation and Cropping:}
Diffusion models often struggle with generating tiny images \cite{croitoru2023diffusion}, whereas PAR images are typically small. To ensure high-quality attribute generation, we first produce high-quality 2784 × 1024 resolution images, then crop them using a pedestrian detector to ensure the pedestrian remains the sole main object.

\begin{figure}[t]
    \centering
    \includegraphics[width=0.85\columnwidth]{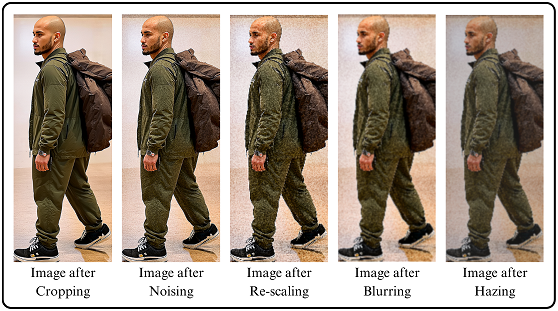}
    \vspace{-2mm}    
    \caption{Image modification as it goes trough the different steps of the proposed degradation process.}
    \label{fig:degradation_process}
    \vspace{-2mm}  
\end{figure}

\vspace{-2mm}
\paragraph{PAR-Style Image Degradation:}
We introduce a degradation module to match PAR image quality, which adds noise, resizes images to simulate pixelation, applies Gaussian blur, and adjusts brightness/contrast to mimic security camera effects in PAR. An example is shown in Figure \ref{fig:degradation_process}. 

\subsection{Integrating Synthetic Data into PAR Datasets}
After generating images for each specific attribute, a protocol is needed to label the new images, integrate them for training, and adjust the loss function for seamless training.

\begin{figure}[t]
    \centering
    \includegraphics[width=0.9\columnwidth]{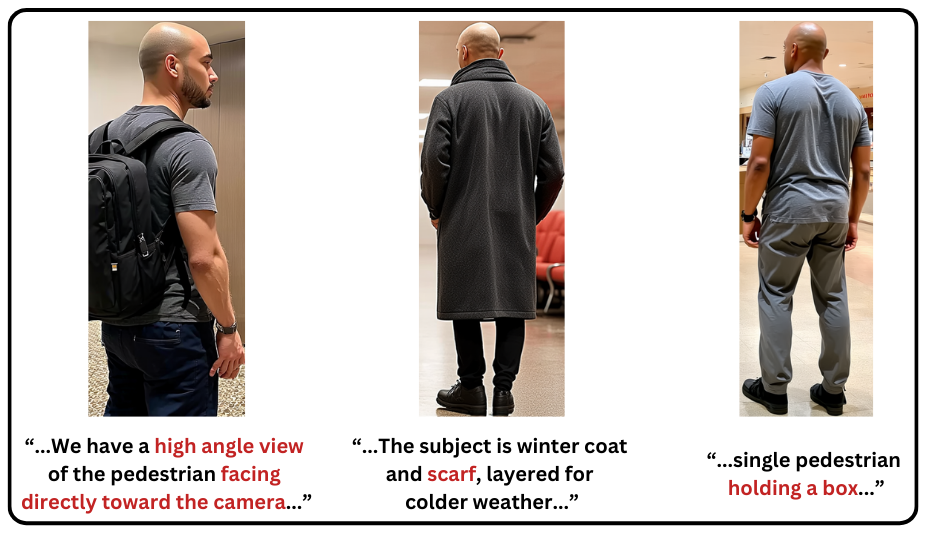}
    \vspace{-3mm}
    \caption{Generated samples with part of the prompt missing on the attributes of the image.}
    \label{fig:missing_attributes}
    \vspace{-5mm}
\end{figure}

\vspace{-3mm}
\paragraph{Automated Data Annotation:}
Manually annotating synthetic images is impractical, so we modify the annotation vectors while preserving their original structure. Instead of one-hot encoding, we introduce a richer labeling scheme: $-1$ (uncertain presence), $1$ (confirmed presence for the target attribute), and $3$ (high confidence based on the prompt). The lack of $0$s (certainty of absence) aligns with dataset inconsistencies where mutually exclusive attributes are not strictly enforced. Similarly, $2$s are avoided to maintain compatibility with existing annotations, where they are later mapped to $1$s in the baseline considered \cite{jia2021rethinking}. The distinction between $1$ (main attribute) and $3$ (other attributes) ensures that only manually verified attributes receive a $1$, as prompts more reliably generate the target attribute when explicitly reinforced at the end, unlike wildcard-based attributes, which may occasionally fail (see Figure \ref{fig:missing_attributes}).

\vspace{-2mm}
\paragraph{Loss Function Modification:}
To employ our augmented images and labels, we introduce \textit{Binary Cross Entropy Augmented} ($\mathcal{L}_{aug}$), defined for each $ith$ train image as:

\vspace{-4mm}
\begin{equation} \label{eq:BCE_augmented}
\begin{aligned}
\mathcal{L}_{aug} = -\frac{1}{M} \sum_{m=1}^{M} w_{i,m} \cdot \Big( & \mathbf{1}_{y_{i,m} \neq 0} \cdot \log \hat{y}_{i,m} + \\[-15pt]
&  \mathbf{1}_{y_{i,m} = 0} \cdot \log (1 - \hat{y}_{i,m}) \Big)
\end{aligned}
\vspace{-2mm}
\end{equation}
where $y_{i,m}, m \in \{0, 1\}^M$ is the (augmented or original) label for the $m$ attribute, $\hat{y}_{i,m}$ is the predicted probability, $M$ is the number of attributes, and $w_{i,m}$ weighs the augmented data to optimize training performance:

\vspace{-2mm}
\begin{equation} \label{eq:weights}
w_{i,m} =
\begin{cases}
0, & \text{if } y_{i,m} = -1 \\
\text{weight\_augmented}, & \text{if } y_{i,m} = 3 \\
1, & \text{if } y_{i,m} \in \{0, 1, 2\} \\
\end{cases},
\end{equation}
where attributes labeled as $-1$ are ignored, and those labeled as $3$ contribute proportionally to $weight\_augmented$.

\section{Experimental Results}
We begin by identifying weak attributes across selected datasets, followed by studying key hyperparameters for generating synthetic images, and training a PAR model with both real and synthetic data. Extended results with additional metrics are available in the supplementary material.
\subsection{Setup}
\vspace{-2mm}
\paragraph{Datasets:} We selected the RAP datasets (RAPv1 and RAPv2 \cite{li2018richly}, and RAPzs \cite{jia2021rethinking}), as they share most attributes while retaining distinct characteristics. 
Our study employs 152K images and 35 common attributes, providing a balanced dataset size and a strong framework for attribute analysis. We use the default train-test partitions.

\vspace{-3mm}
\paragraph{Evaluation Metrics:} 
We used the F1 Score, standard in PAR \cite{wang2022pedestrian}, as the harmonic mean of Precision and Recall. 

\vspace{-3mm}
\paragraph{Implementation Details:}Our generative pipeline is built on Stable Diffusion 3 implemented with ComfyUI\footnote{\url{https://github.com/comfyanonymous/ComfyUI}} to handle the integration of the pipeline blocks. For training, we employed a widely used PAR architecture that combines a ResNet backbone with an MLP classifier \cite{jia2021rethinking, kirillova2022visual, galiyawala2021dsa}, being trained with 256x128 images as in \cite{jia2021rethinking,zhou2025heterogeneous}. The code implementation of our proposal is available at Github\footnote{\url{https://github.com/vpulab/PARtxt2img}}.

\subsection{Assessing the Attributes to Improve}
\label{sec:results-identifying-attributes}
\paragraph{Optimizing PAR Model Training:}
To ensure model robustness, we optimized weight decay (WD) and batch size (BS) while keeping other hyperparameters as in \cite{jia2021rethinking}: optimizer (Adam), learning rate ($1e^{-3}$) and 30 epochs. We analyzed 16 configurations by combining four WD ($5e^{-5}$ to $5e^{-2}$) and four BS (8 to 64), finding BS=16 and WD=$5e^{-3}$ to yield the best results for all datasets (see Table \ref{tab:optimized_baseline}).

\begin{table}[t]
\centering
\caption{Performance comparison of the baseline and optimized model on the RAP datasets. Best results in bold.}
\vspace{-2mm}
\resizebox{\columnwidth}{!}{
\begin{tabular}{c|cc|c|c|c|}
\hline
 \textbf{PAR} & \multicolumn{2}{c|}{\textbf{Hyperparams}} & \textbf{RAPv1} & \textbf{RAPv2} & \textbf{RAPzs} \\
 \cline{2-6}
 \textbf{Approach} & \textbf{WD} & \textbf{BS}  & \textbf{F1} & \textbf{F1} & \textbf{F1} \\
\hline
\hline
Baseline \cite{jia2021rethinking} & $5e^{-4}$ & 64 & 79.95 & 78.68 & 77.66\\
Optimized & $5e^{-3}$ & 16 &\textbf{80.30} & \textbf{79.38} & \textbf{78.20}  \\
\hline
\end{tabular}
}
\label{tab:optimized_baseline}
\end{table}

\begin{table}[t]
\centering
\captionsetup{justification=centering}
\caption{Top-10 lowest-performing attributes in RAP datasets based on the proposed scoring methodology. Action and 'Others' attributes are excluded from this study.}
\vspace{-2mm}
\resizebox{\columnwidth}{!}{%
\begin{tabular}{c|c|c|c}
\hline
\textbf{Rank}   & \textbf{RAPv1}    & \textbf{RAPv2}    & \textbf{RAPzs} \\ 
\hline \hline
1               & hs-BaldHead       & ub-Others         & hs-BaldHead    \\ \hline
2               & hs-Muffler        & shoes-Cloth       & ub-ShortSleeve \\ \hline
3               & ub-Tight          & lb-ShortSkirt     & shoes-Cloth    \\ \hline
4               & attach-PaperBag   & lb-Dress          & lb-ShortSkirt  \\ \hline
5               & hs-Hat            & attach-HandBag    & attach-PlasticBag \\ \hline
6               & ub-SuitUp         & attach-PlasticBag & attach-PaperBag \\ \hline
7               & ub-ShortSleeve    & attach-PaperBag   & ub-SuitUp \\ \hline
8               & lb-ShortSkirt     & hs-BaldHead       & ub-Tight \\ \hline
9               & AgeLess16         & hs-Hat            & attach-Backpack \\ \hline
10              & BodyFat           & AgeLess16         & AgeLess16  \\ \hline
\end{tabular}
}
\label{table:worse_attributes}
\end{table}

\vspace{-3mm}
\paragraph{Attribute Ranking and Selection:}
Table \ref{table:worse_attributes} shows the lowest-ranked attributes for the proposed scores in Sec. \ref{sec:weak-attributes}, many of which suffer from class imbalance, ambiguity, or occlusion. Given the difficulty of replicating actions and vague attributes like 'others' using text-guided synthetic images, we focus on attributes easier to describe, and thus more reliable to synthesize. We prioritize augmentation for \textit{hs-BaldHead}, \textit{lb-ShortSkirt}, \textit{AgeLess16}, \textit{ub-SuitUp}, and \textit{attach-PaperBag}, as they frequently appear in datasets.

\setcounter{table}{5}
\begin{table*}[ht]
\setlength{\tabcolsep}{3pt}
\centering
\caption{Attribute-level F1-score comparison for different data augmentation (DA) percentages on the RAP datasets. Best results for each dataset and attribute are in bold.'*' refers to our optimized model (see Table \ref{tab:optimized_baseline}).}
\label{tab:attribute_F1_results}
\vspace{-2mm}
\resizebox{1\textwidth}{!}{%
\begin{tabular}{c|ccc|ccc|ccc|ccc|ccc|}
\hline
\textbf{DA} & \multicolumn{3}{|c}{\textbf{\textit{hs-BaldHead}}} & \multicolumn{3}{|c}{\textbf{\textit{lb-ShortSkirt}}} & \multicolumn{3}{|c}{\textbf{\textit{AgeLess16}}} &  \multicolumn{3}{|c}{\textbf{\textit{ub-SuitUp}}} & \multicolumn{3}{|c|}{\textbf{\textit{attach-PaperBag}}} \\ \cline{2-16}
\textbf{(\%)}  & \textbf{RAPv1} & \textbf{RAPv2} & \textbf{RAPzs} & \textbf{RAPv1} & \textbf{RAPv2} & \textbf{RAPzs} & \textbf{RAPv1} & \textbf{RAPv2} & \textbf{RAPzs} & \textbf{RAPv1} & \textbf{RAPv2} & \textbf{RAPzs} & \textbf{RAPv1} & \textbf{RAPv2} & \textbf{RAPzs} \\ \hline \hline
0*  & 47.46 & 66.67 & 37.50 & \textbf{68.62} & \textbf{62.54} & \textbf{51.30} & 75.82 & 70.14 & \textbf{59.02} & \textbf{69.29} & 65.83 & 66.67 & 45.79 & 42.90 & 9.30\\
\hline
50  & 55.88 & 63.86 & 28.57 & 65.57 & 61.65 & 49.43 & 75.86 & 71.64 & 44.68 & 67.70 & 65.50 & 64.25& 51.35 & 41.48 & 17.72\\    
100 & 60.61 & \textbf{68.24} & 37.50 & 68.51 & 59.35 & 46.56 & 76.82 & 71.77 & 48.00 & 68.42 & \textbf{66.87} & 67.94 & 51.10 & 41.09 &20.93\\
200 & 61.54 & 66.13 & 26.67 & 66.11 & 59.54 & 50.00 & 77.12 & 72.11 & 56.07 & 67.02 & 64.55 & 63.68 & 49.54 & 40.00 & 25.88\\
300 & 59.70 & 67.04 & 36.36 & 65.50 & 59.72 & 45.53 & 78.05 & \textbf{72.83} & 58.72 & 66.99 & 65.07 & 66.02 & \textbf{54.17} & 41.64 & \textbf{33.33}\\
400 & 61.54 & 64.13 & 36.36 & 67.68 & 61.72 & 45.42 & \textbf{78.67} & 72.25 & 52.94 & 68.25& 66.00 & 65.37 & 45.37 & 44.14 &30.93\\
500 & \textbf{63.49} & 68.16 & \textbf{40.00} & 63.49 & 56.60 & 48.28 & 76.19 & 72.14 & 56.41 & 66.30 & 63.19 & \textbf{68.32} & 49.28 & \textbf{45.14} & 23.38\\ 
\hline
\end{tabular}
}
\vspace{-5mm}
\end{table*}

\setcounter{table}{3} 

\subsection{Impact of Synthetic Data Augmentation}
\paragraph{Parameter Selection for Generation:}
To optimize augmentation, we selected the weakest attribute \textit{hs-BaldHead} identified in the subsection \ref{sec:results-identifying-attributes}, and we evaluated two key parameters on the RAP datasets: image noise level (see Sec. \ref{sec:generation-pipeline}) and augmented weight (see Eq. \ref{eq:weights}). By default, each real image was augmented five times, using a medium noise level and 0.5 augmentation weight. We tested three noise blending levels (25\%, 50\%, 75\%) in Table \ref{tab:BaldHead-noise}, with medium noise performing best. Augmented weight, explored from 0 to 1 with four values, was optimal at 0.5 (see Table \ref{tab:BaldHead-weight}). These best values were used for all subsequent experiments.

\begin{table}[t]
\centering
\caption{Attribute-level F1-score comparison of noise levels for \textit{hs-BaldHead} attribute on the RAP datasets. An augmentation weight of 0.5 was employed.}
\vspace{-2mm}
\label{tab:BaldHead-noise}
\begin{tabular}{l|c|c|c}
\hline
\textbf{Noise level} & \textbf{RAPv1} & \textbf{RAPv2} & \textbf{RAPzs}\\
\hline
\hline
Big noise (75\%) & 53.12 & 66.67 & 23.53\\
Medium noise (50\%) & \textbf{63.49} & \textbf{68.16} & \textbf{40.00}\\
Small noise (25\%)& 58.62 & 66.29 & \textbf{40.00}\\
\hline
Optimized from Table \ref{tab:optimized_baseline} & 47.46 & 66.67 & 37.50 \\
\hline\end{tabular}
\vspace{-2mm}
\end{table}

\begin{table}[t]
\centering
\caption{Attribute-level F1-score comparison of augmentation weights for \textit{hs-BaldHead} attribute on the RAP datasets. A medium noise level was employed.}
\vspace{-2mm}
\label{tab:BaldHead-weight}
\resizebox{\columnwidth}{!}{%
\begin{tabular}{c|c|c|c}
\hline
\textbf{Augmentation Weight} & \textbf{RAPv1}  & \textbf{RAPv2}  & \textbf{RAPzs}\\
\hline
\hline
0.25 & 61.29 & 65.17 & 14.29 \\
0.5 & \textbf{63.49} & \textbf{68.16} & \textbf{40.00} \\
0.75 & 53.33 & 64.77 & \textbf{40.00} \\
 1 & 54.24 & 65.90 & 18.18 \\
\hline
Optimized from Table \ref{tab:optimized_baseline} & 47.46 & 66.67 & 37.50 \\
\hline
\end{tabular}
}
\vspace{-2mm}
\end{table}

\vspace{-2mm}
\paragraph{Per-Attribute Performance Gains:}
We analyzed attribute-level F1-scores across augmentation levels (0.5–5 images per sample), ensuring consistency by using incrementally overlapping sets (e.g., 100\% includes 50\%). Table \ref{tab:attribute_F1_results} shows that most attributes see gains as augmentation increases, confirming the effectiveness of synthetic data in enhancing PAR. Across datasets, RAPv2 benefits the most where RAPzs remains challenging, highlighting the need for further adaptation strategies in zero-shot settings. While 500\% provides the best F1-score in RAPv1 for \textit{hs-BaldHead}, other attributes (e.g., \textit{lb-ShortSkirt}, \textit{attach-PaperBag}) do not consistently improve with extreme augmentation, indicating potential overfitting.

\vspace{-2mm}
\paragraph{Overall Performance:}
After optimizing augmentation per attribute, we tested simultaneous augmentation across datasets. We evaluated two models: one using 500\% augmentation for all attributes and another applying each attribute's optimal augmentation level. Since \textit{lb-ShortSkirt} showed no improvement, we also tested augmentation with only \textit{hs-BaldHead} and \textit{AgeLess16}. 
Table \ref{tab:final_results} compares state-of-the-art image augmentation methods, averaging results over 10 runs using the same architecture from \cite{jia2021rethinking}. It can be seen that our approach improves overall performance by augmenting 2–3 of 52 attributes, as compared to the baseline (i.e., Horizontal Flip \cite{jia2021rethinking}). Our method performs best on RAPv1 and RAPzs, ranking among the top 2 across all datasets and highlighting its robustness for the datasets considered. Interestingly, including \textit{lb-ShortSkirt} often yielded the best or highly competitive results, suggesting an indirect boost to other attributes.

\setcounter{table}{6}
\begin{table}[t]
\setlength{\tabcolsep}{3pt}
\centering
\caption{F1-score comparison of selected approaches for Image Data Augmentation (DA) on RAP datasets. Best, second-, and third-best results are shown in \textcolor{blue}{blue}, \textcolor{green}{green}, and \textcolor{red}{red}, respectively. Selected DA approaches perform augmentation for all attributes available in each dataset, whereas our proposal only focuses on the attributes \textit{hs-BaldHead} (BH), \textit{lb-ShortSkirt} (SK), \textit{AgeLess16} (A16)}
\vspace{-2mm}
\label{tab:final_results}
\resizebox{\columnwidth}{!}{%
\begin{tabular}{|c|ccc|ccc|}
\hline
 \textbf{DA approach} & \textbf{BH} & \textbf{SK} & \textbf{A16} & \textbf{RAPv1}  & \textbf{RAPv2}  & \textbf{RAPzs}\\
\hline
\hline
Proposed - 500\% & \checkmark & - & \checkmark & \textcolor{red}{80.51} & 79.59 & 78.16\\
Proposed - 500\% & \checkmark & \checkmark & \checkmark  & \textcolor{blue}{80.60}  & 79.67 & \textcolor{blue}{78.62} \\ 
Proposed - Best \% & \checkmark & - & \checkmark & 80.38 & \textcolor{green}{79.77} & \textcolor{red}{78.26}\\
Proposed - Best \% & \checkmark & \checkmark & \checkmark  & 80.50 & \textcolor{red}{79.76} & \textcolor{green}{78.52}\\ 
\hline
Horizont. Flip \cite{jia2021rethinking} & - & - &- & 79.95 & 78.68 & 77.66 \\ 
GANs \cite{fabbri2017generative} & - & - &- & 77.83 & - & - \\ 
AutoAug \cite{cubuk2019autoaugment} & - & - &- & 80.39 & 79.54 & 78.25 \\
MixUp \cite{zhang2017mixup}  & - & - &- & 80.35 & \textcolor{blue}{80.11} & 77.57 \\
RandAug \cite{cubuk2020randaugment}  & - & - &- & 80.26 & 79.72 & 77.34 \\
TrivAug \cite{muller2021trivialaugment}  & - & - &- & \textcolor{green}{80.55} & 79.64 & 77.46 \\
AugMix \cite{hendrycks2019augmix}  & - & - &- & 80.42 & 79.44 & 77.19 \\
\hline

\end{tabular}
}
\end{table}

\section{Conclusions and Future Work}
This work introduced a data-centric approach for improving Pedestrian Attribute Recognition by augmenting training data with synthetic images generated by diffusion models. Instead of increasing model complexity, we identified weakly recognized attributes and selectively enhanced their representation. Experimental results demonstrated consistent improvements in attribute recognition, particularly for underrepresented ones, while maintaining overall model robustness. Notably, our method achieved comparable or superior results to more complex strategies, highlighting the effectiveness of targeted synthetic augmentation. 

Future work will extend our proposal to more attributes, architectures, and datasets while exploring methods for automatically estimating the optimal augmentation size.

\section*{Acknowledgments}
This work has been partially supported by the Spanish Government (PID2021-125051OB-I00) and is part of preliminary work in project TEC-2024/COM-322 (IDEALCV-CM) funded by the Regional Government of Madrid. 
The authors would also like to thank the anonymous reviewers for the valuable feedback provided.

{\small
\bibliographystyle{bib/ieee_mine}
\bibliography{bib/egbib}
}

\clearpage
\appendix
\onecolumn
\begin{center}
\large{\bf{SUPPLEMENTARY MATERIAL}}    
\end{center}

This supplementary material offers additional details on the prompt definition of our proposal in Section \ref{sec:input-prompts}. Moreover, we present extended information for the experiments of the paper regarding the baseline optimization (Section \ref{sec:extended-experiments-optimization}), attributes analysis and ranking for the chosen datasets (Section \ref{sec:extended-experiments-attributes}), and synthetic data augmentation and training (Section \ref{sec:extended-experiments-dataaugmentation}). Finally, we provide configuration details for the ComfyUI tool employed for experimentation in Section \ref{annex:ComfyUIDetails}.

\section{Input Prompts for Synthetic PAR-like Image Generation with attribute variation} 
\label{sec:input-prompts}
After identifying the attributes to improve, our pipeline based on text-to-image generation generates synthetic images based on prompts. Here we provide an extended description of prompts employed.

\vspace{-2mm}
\subsection{Prompt structure} 
\begin{figure}[b!]
    \centering
    \includegraphics[width=0.70\columnwidth]{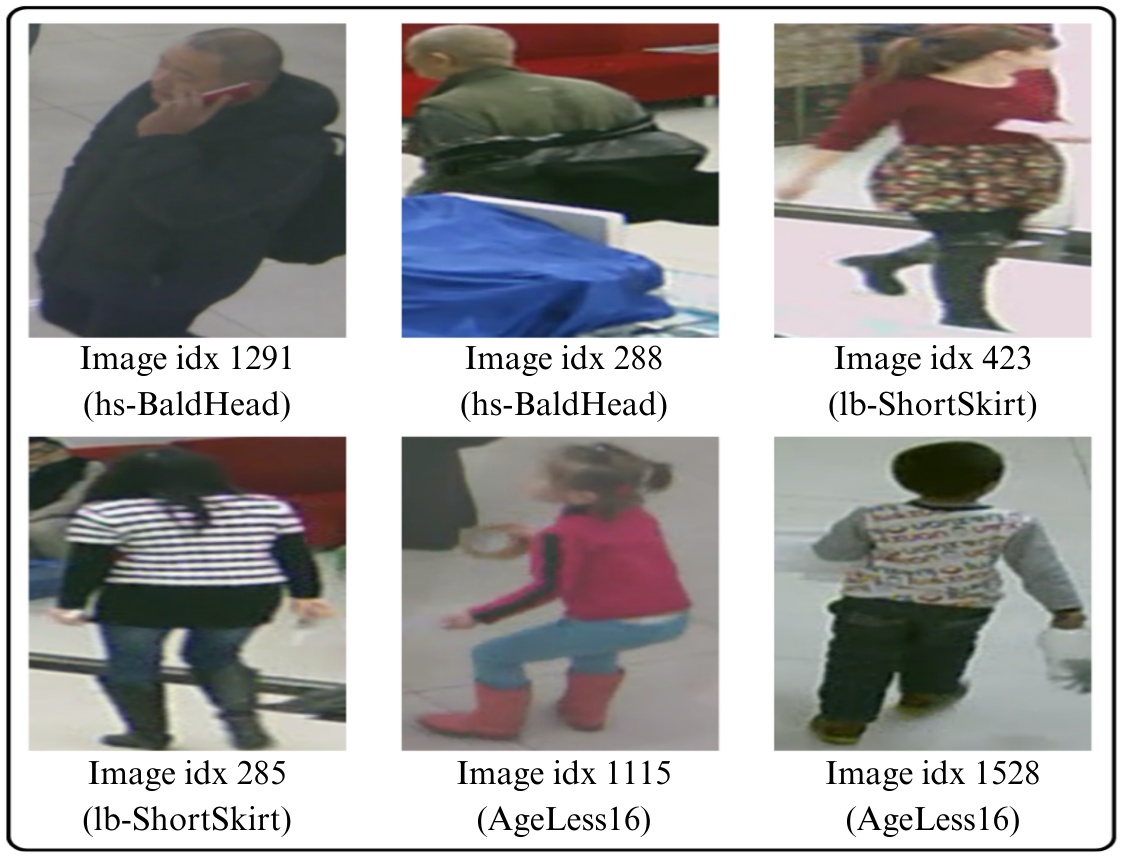}
    \caption{Dataset ground-truth samples}
    \label{fig:dataset}
    \vspace{-2mm}
\end{figure}

Designing an effective prompt is critical for obtaining robust results that accurately reflect the underlying dataset while providing sufficient variability to avoid overfitting to specific characteristics. In our work, the generation process must produce images resembling those in the RAP  datasets, specifically depicting pedestrians cropped within airport environments and performing diverse actions as captured by security cameras. This inherently results in samples with low image quality. Examples from the original dataset are shown in Figure \ref{fig:dataset}.

\begin{figure}[t]
    \centering
    \includegraphics[width=0.70\columnwidth]{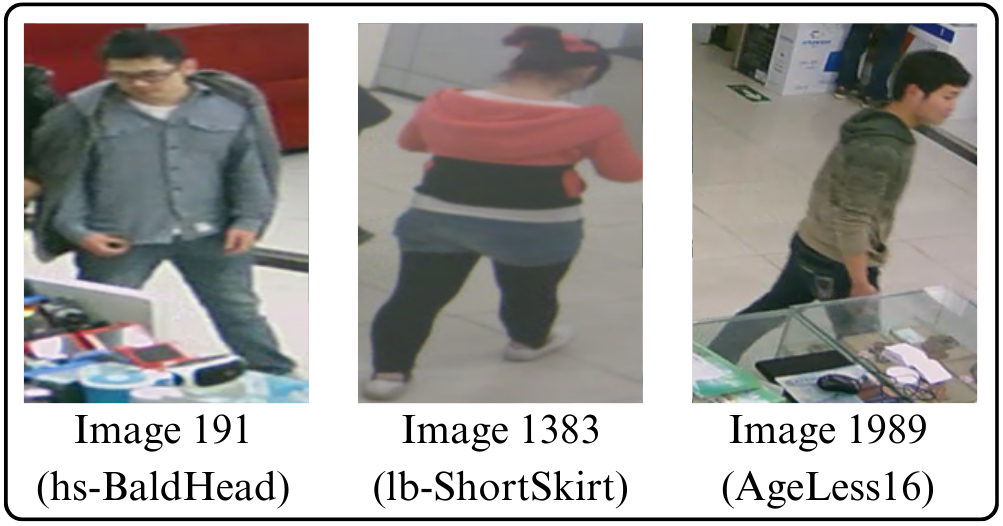}
    \caption{Noisy dataset ground-truth samples}
    \label{fig:dataset_noise}
    \vspace{-2mm}
\end{figure}

It is important to note, however, that upon examining the dataset, it is not difficult to find examples where the labels are incorrectly assigned (Figure \ref{fig:dataset_noise}). In our case, the goal is to align with the dataset while avoiding the introduction of noise. For \textit{hs-BaldHead}, at least part of the scalp must be visible; for \textit{lb-ShortSkirt}, the garment must be open between the legs and end between the waist and the knee; and for \textit{AgeLess16}, the pedestrian must appear to fall within the indicated age range.

\begin{figure}[t]
    \centering
    \includegraphics[width=0.90\columnwidth]{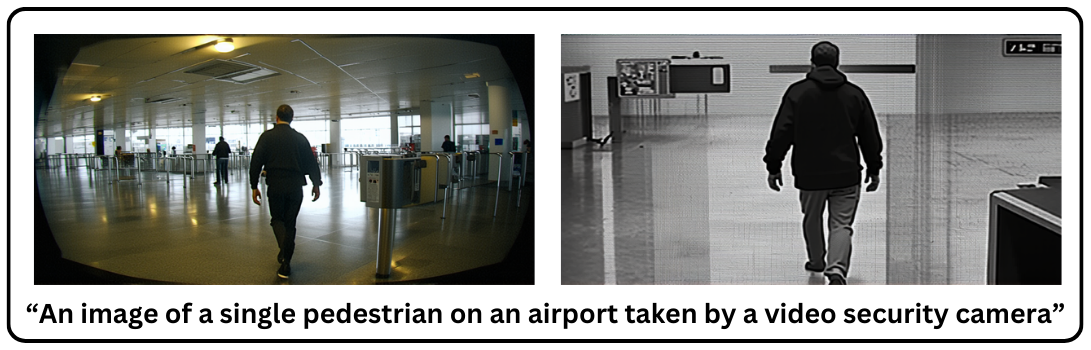}
    \caption{Samples generated with the given prompt using keywords such as "security camera"}
    \label{fig:misgenerations}
\end{figure}

In our preliminary experiments, generating images with content similar to the target dataset was relatively straightforward using varied prompts. However, replicating the exact image quality proved challenging when relying solely on prompt-based generation. For instance, including descriptors such as “security camera” often led the model to produce images that deviated from the target dataset in terms of tonal distribution, watermark artifacts, noise patterns, and even fisheye distortions (see Figure \ref{fig:misgenerations}). Furthermore, the use of negative prompts did not sufficiently alleviate these discrepancies.

As a result, we first generate high-definition images (leveraging the model's superior performance in this task) and subsequently adjust their visual characteristics to align with the target dataset. Even taking advantage of not restraining quality by prompting, and although diffusion models excel generating images with considerable variability, a static prompt is inadequate to capture the full spectrum of variations inherent in the dataset. 

To address this limitation, our approach follows the \textit{Wildcards and Template-Based Prompts} strategy, where wildcards are dynamically replaced with pre-configured phrases based on predefined probabilities. This allows for richer, more complex prompts that better reflect dataset characteristics. We analyzed PAR datasets to define each wildcard, resulting in the positive prompt shown in Figure \ref{fig:main_promt}. Additionally, negative prompts were used to exclude undesired concepts. Thus, the resulting prompts provide a richer variability and complexity (see examples in Figure \ref{fig:generated_samples}).

\vspace{5mm}

\noindent\fbox{%
    \parbox{\linewidth}{%
        A highly detailed, ultra high definition image of a single pedestrian \textbf{\_\_poses\_\_} on \textbf{\_\_backgrounds\_\_}, with the pedestrian fully visible and centered in the frame. The pedestrian is visible from head to toe and is the primary focus, and the scene features realistic, natural colors. We have \textbf{\_\_views\_\_}. The subject is wearing a \textbf{\_\_styles\_\_}. \textbf{\_\_colors\_\_}. It can also be seen how the pedestrian \textbf{\_\_attributes\_\_}.
    }%
}
\vspace{-2mm}
\captionof{figure}{Positive prompt for PAR-like image generation.}
\label{fig:main_promt}
\setlength{\parindent}{1pc} 

\begin{figure}[t]
    \centering
    \includegraphics[width=0.65\columnwidth]{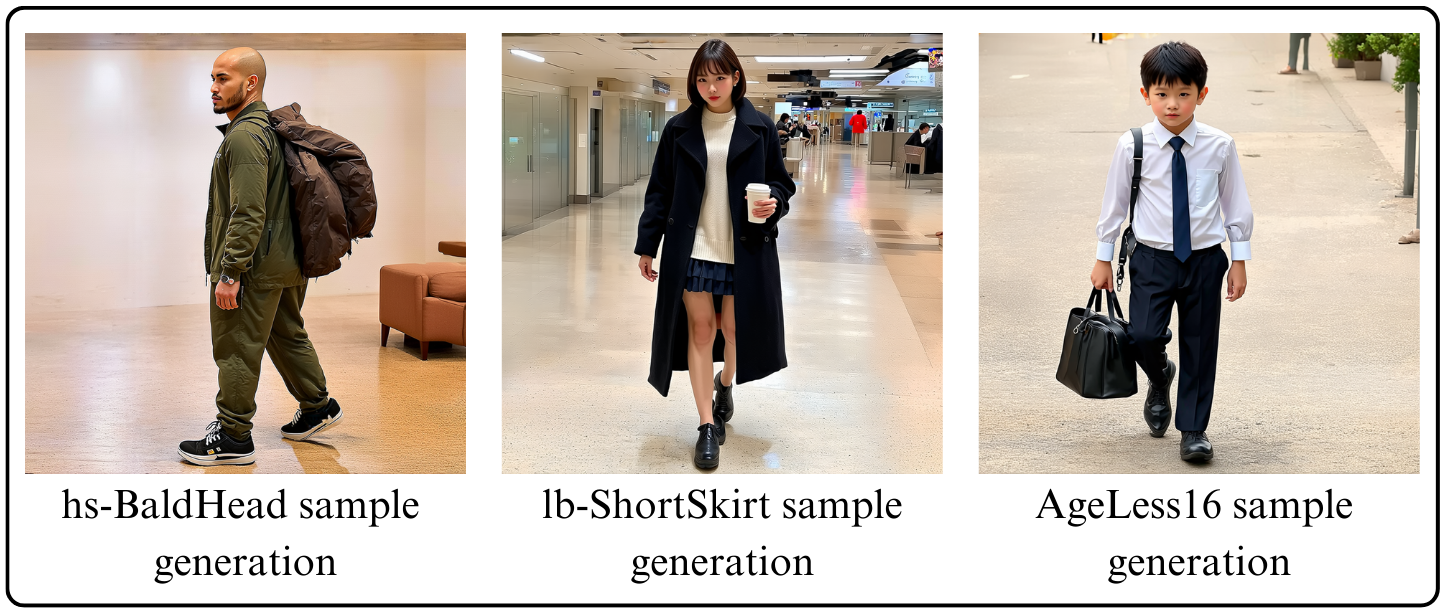}
    \caption{Samples generated with the proposed prompt for each one of the pedestrian attributes.}
    \label{fig:generated_samples}
\end{figure}

\subsection{Wildcard definition} 

To characterize each wildcard, we performed a comprehensive analysis of the dataset to accurately mirror its inherent conditions. In the following, we detail the definition of each wildcard and provide representative examples. These examples, documented in Subsection \ref{sec:prompt-definition}, are uniformly applied during practical implementation.

\begin{itemize} 
    \item \textbf{Poses}: To prevent pedestrians from consistently appearing in a walking stance, “poses” refer to diverse postures (e.g., “crossing arms while standing”, “moving with hands behind the back”) or subtle actions (e.g., “using a smartphone”, “holding paper or documents”) observed directly in the dataset. 
    \item \textbf{Backgrounds}: Although all images are set in airport interiors, the backgrounds vary. Based on observed examples, wildcards such as “an airport terminal waiting area” or “an indoor shopping center walkway” are introduced. Some descriptions, despite not explicitly mentioning the airport, mimic backgrounds present in the dataset. 
    \item \textbf{Views}: Views adjust the visible portion of the pedestrian to enhance generalization. Examples include “a clear view of the back of the person, captured from behind” or “a high-angle view of the profile of the pedestrian”, replicating typical security camera angles. 
    \item \textbf{Styles}: In generating images for the bald-head attribute, we observed that the model tended to depict bald individuals wearing either suits or gym attire when representing people of color. To introduce variability, clothing styles such as “casual t-shirt and jeans” or “winter coat and scarf, layered for colder weather” were incorporated. 
    \item \textbf{Colors}: This wildcard pertains to the color scheme of the previously mentioned clothing styles. Examples include descriptions like “practical and functional colors, with dark tones such as black, grey, and navy” or “vibrant, eye-catching colors, including bright reds, oranges, and yellows.” 
    \item \textbf{Attributes}: This wildcard, expressed in a separate sentence from the others, specifically addresses the augmented attribute. To ensure consistency with the dataset, the term “asian” is incorporated, yielding examples such as “is a bald Asian person”, “is an Asian person wearing a short skirt with tights or leggings”, and “is an Asian kid or teenager clearly not older than 15 years old.” 
\end{itemize}

\subsection{Prompt definition} 
\label{sec:prompt-definition}

The prompts used for image generation are described below, including both positive and negative prompts, as well as the wildcards employed. The objective is to document exactly how each image could have been generated. \\

First, the negative prompt used to guide the generation was: \\

\texttt{"bad quality, poor quality, Partial figures, cropped bodies, cut-off limbs, headless or footless pedestrians, close-up shots, extreme zoom, obscured views, hidden or partially visible subjects, cropped at the knees or waist, off-frame figures, incomplete visibility, overly zoomed-in perspectives"} \\

This aims, in general, to produce images of the highest possible quality where a pedestrian is clearly visible as the main element of the image. Additionally, for the generation of \textit{shortskirts}, the following was appended to the prompt to avoid generations resembling short skirts that do not meet the defined criteria: \\

\texttt{", no pants, no long skirts, no oversized clothing."} \\

Regarding the positive prompt, the one used is as mentioned in the corresponding section of the report, specifically: \\

\texttt{"A highly detailed, ultra high definition image of a single pedestrian \_\_poses\_\_ on \_\_backgrounds\_\_, with the pedestrian fully visible and centered in the frame. The pedestrian is visible from head to toe and is the primary focus, and the scene features realistic, natural colors. We have \_\_views\_\_. The subject is wearing a \_\_styles\_\_. \_\_colors\_\_. It can also be seen how the pedestrian 
\_\_attributes\_\_."} \\

As for the wildcards and their intended purposes, below are the specific values they can contain: \\
~\newline
\_\_attributes\_\_: (\textit{One of these is selected based on the attribute being augmented})
\begin{itemize}
    \setlength{\itemsep}{0pt}
    \item is a bald asian person
    \item is an asian person wearing a short skirt with tights or leggings
    \item is an asian kid or teenager clearly not older than 15 years old
    \item is a female asian child
\end{itemize}

~\newline
\_\_backgrounds\_\_:
\begin{itemize}
    \setlength{\itemsep}{0pt}
    \item a shopping mall interior with glass storefronts
    \item a shopping center walkway
    \item an indoor shopping mall food court
    \item an indoor escalator near shops
    \item an indoor walkway near escalators
    \item an indoor office building corridor
    \item an office lobby with elevators
    \item an indoor public lobby with seating
    \item an airport terminal waiting area
    \item an airport security checkpoint
\end{itemize} 
~\newline
\_\_colors\_\_:
\begin{itemize}
    \setlength{\itemsep}{0pt}
    \item The outfit is vibrant and full of bold, eye-catching colors, such as bright reds, oranges, and yellows.
    \item The look incorporates a balanced mix of soft and bright tones, like pastel pinks, light blues, and subtle greens, creating a modern, trendy appearance.
    \item The style is composed of subtle, muted shades with minimal color contrast, such as grey, beige, and light brown, blending naturally into the background.
    \item The outfit follows a monochromatic palette, staying within a single color family, like different shades of black, white, or navy, giving the pedestrian a sleek, cohesive look.
    \item The style features natural, earthy hues that create a calm, understated appearance, with tones like olive green, brown, and khaki.
    \item The colors are practical and functional, with dark tones like black, grey, and navy.
    \item The outfit contrasts between dark and light elements, with bright whites or light greys paired with deep blues or blacks, creating a visually striking yet balanced style.
    \item The pedestrian's clothing includes soft neutral tones like beige, off-white, and light grey, creating a subtle, relaxed appearance.
    \item The outfit incorporates rich, formal colors like maroon, charcoal grey, and dark green.
    \item The style is colorful yet refined, using contrasting accents like a red accessory paired with a neutral-toned outfit, drawing attention while maintaining sophistication.
\end{itemize}
\_\_poses\_\_:
\begin{itemize}
    \setlength{\itemsep}{0pt}
    \item walking
    \item standing still
    \item using a smartphone
    \item holding a shopping bag
    \item carrying a backpack
    \item carrying a shoulder bag
    \item carrying a small bag in the front
    \item carrying a small bag on the side
    \item holding a shopping bag with both hands
    \item with their hands in the pockets
    \item crossing arms while standing
    \item adjusting clothing
    \item looking to the side
    \item looking down at a phone
    \item walking while holding a coffee cup\#(action-Holding)
    \item standing with one hand on hip
    \item holding a jacket
    \item carrying a jacket over the shoulder
    \item holding a water bottle
    \item carrying multiple items in hand
    \item moving with hands behind the back
    \item holding paper or documents
    \item holding a box
    \item checking or exploring a box
    \item carrying a box
    \item moving a big box or large items
    \item carrying multiple bags
\end{itemize}
~\newline
\_\_styles\_\_:
\begin{itemize}
    \setlength{\itemsep}{0pt}
    \item casual t-shirt and jeans
    \item formal business suit with a tie
    \item business casual outfit with a button-up shirt and slacks
    \item jacket over a t-shirt with jeans
    \item hoodie and sweatpants
    \item sporty outfit with a windbreaker and jogging pants
    \item leather jacket, streetwear style and jeans, streetwear style
    \item long overcoat over a sweater and slacks
    \item short-sleeved shirt with chinos
    \item winter coat and scarf, layered for colder weather
    \item light jacket, appropriate for indoor walking and jeans, appropriate for indoor walking
    \item blazer with a button-up shirt and chinos
    \item polo shirt, semi-casual look with khaki pants, semi-casual look
    \item denim jacket with a t-shirt and sneakers
    \item sweater over a collared shirt with dress pants
    \item windbreaker, outdoor casual style and jeans, outdoor casual style
    \item puffer jacket, typical for colder weather with casual pants, typical for colder weather
    \item cardigan over a long-sleeve shirt and jeans
\end{itemize}

In the case of generating shortskirts, all attributes related to pants have been excluded, leaving the following:
\begin{itemize}
    \setlength{\itemsep}{0pt}
    \item casual t-shirt
    \item formal business suit with a tie
    \item business casual outfit with a button-up shirt
    \item jacket over a t-shirt
    \item hoodie
    \item casual dress or skirt with a jacket
    \item sporty outfit with a windbreaker
    \item leather jacket, streetwear style
    \item long overcoat over a sweater
    \item short-sleeved shirt
    \item winter coat and scarf, layered for colder weather
    \item light jacket, appropriate for indoor walking
    \item blazer with a button-up shirt
    \item polo shirt, semi-casual look
    \item denim jacket with a t-shirt and sneakers
    \item summer dress with sandals
    \item sweater over a collared shirt
    \item windbreaker, outdoor casual style
    \item puffer jacket, typical for colder weather
    \item cardigan over a long-sleeve shirt
\end{itemize}
~\newline
\_\_views\_\_:
\begin{itemize}
    \setlength{\itemsep}{0pt}
    \item a clear front view of the subject, facing directly toward the camera
    \item a complete side view of the subject, captured in profile
    \item a clear view of the back of the person, captured from behind
    \item a high angle view of the pedestrian facing directly toward the camera
    \item a high angle view of the profile of the pedestrian
    \item a high angle view of the back of the pedestrian, as the image is captured from behind
\end{itemize}

\newpage 
\section{Extended experimental results for optimizing PAR Model Training}
\label{sec:extended-experiments-optimization}
The architecture and training employed follows the general structure provided by Jia et al. \citesupp{jia2021rethinking}. However, to ensure that the attribute analysis is as meaningful as possible, the model has undergone a hyperparameter optimization per dataset before obtaining the baseline results, so that the model analyzed has achieved the best performance.

\paragraph{Optimization for RAPv1:}
The optimization process commenced with RAPv1, the smallest subset within the RAP datasets. For efficiency, a set of relevant hyperparameters was selected to tune using a one-dimensional grid search, addressing each parameter individually. More complex search strategies were discarded due to the large number of hyperparameters involved.

The chosen hyperparameters were those that significantly influenced the model's performance or that differed with the baseline \citesupp{jia2021rethinking} and the latest GitHub implementation. Based on these criteria, the final set included batch size, classifier, learning rate, optimizer, and weight decay.
Tables \ref{tab:RAP1_bald_weightDecay} and \ref{tab:RAP1_bald_batchSize} summarize the parameters that, when adjusted, led to superior performance compared to the baseline reported by the original authors.

\begin{table}[t]
\centering
\captionsetup{justification=centering}
\caption{Performance metrics for different weight decay values (RAPv1).}
\label{tab:RAP1_bald_weightDecay}
\vspace{-4mm}
\begin{tabular}{c|c|c|c|c|c}
\hline
\textbf{Weight decay} & \textbf{ma} & \textbf{Acc} & \textbf{Prec} & \textbf{Rec} & \textbf{F1} \\ 
\hline \hline
5e-2            & 68.77  & 65.03  & \textit{80.87}     & 75.16  & 77.91  \\ \hline
5e-3            & 78.14  & \textbf{\textit{68.97}}  & \textbf{\textit{81.2}}  & \textit{80.13} & \textbf{\textit{80.48}} \\ \hline
5e-4            & \textbf{\textit{79.76}} & \textit{68.47}  & 80.11  & \textbf{\textit{80.46}}  & \textit{80.29} \\ \hline
5e-5            & 78.71  & 67.82  & 80.15  & 79.52  & 79.84  \\ \hline \hline
5e-4 (theirs)   & 79.27  & 67.98  & 80.19  & 79.71  & 79.95  \\ \hline
\end{tabular}
\end{table}

\begin{table}[t]
\centering
\captionsetup{justification=centering}
\caption{Performance metrics for different batch sizes (RAPv1).}
\label{tab:RAP1_bald_batchSize}
\vspace{-2mm}
\begin{tabular}{c|c|c|c|c|c}
\hline
\textbf{Batch size} & \textbf{ma} & \textbf{Acc} & \textbf{Prec} & \textbf{Rec} & \textbf{F1} \\
\hline\hline
8          & \textbf{80.86} &  68.36 & 79.32 & \textbf{ 81.22} &  80.26 \\
\hline
16         &  80.78 & \textbf{ 68.68} & 79.98 &  80.99 & \textbf{ 80.66} \\
\hline
32         &  79.76 &  68.47 & 80.11 &  80.46 &  80.29 \\
\hline
64         & 78.88 & 67.94 & 80.18 & 79.59 & 79.88 \\
\hline\hline
64 (theirs) & 79.27 & 67.98 & \textbf{80.19} & 79.71 & 79.95 \\
\hline
\end{tabular}%
\end{table}

As shown in Table \ref{tab:RAP1_bald_weightDecayBatchSize_combination}, we improved the overall results by individually modifying weight decay and batch size. Based on these findings, we evaluated a combined configuration of the best-performing settings for both parameters. However, jointly modifying the parameters yielded inferior results, leading us to adjust only the batch size, reducing it to 16. The final model was executed multiple times to ensure robust performance, and the final optimized results on RAP1 are reported in Table \ref{tab:RAP1_bald_optimized}.

\begin{table}[t]
\centering
\captionsetup{justification=centering}
\caption{Performance metrics comparison with best batch size, weight decay and combination (RAPv1).}
\label{tab:RAP1_bald_weightDecayBatchSize_combination}
\vspace{-2mm}
\begin{tabular}{c|c|c|c|c|c}
\hline
\textbf{Modification} & \textbf{ma} & \textbf{Acc} & \textbf{Prec} & \textbf{Rec} & \textbf{F1} \\ 
\hline \hline
WD 5e-3             & 78.14     & \textbf{68.97} & \textbf{81.2} & 80.13     & 80.48 \\ \hline
BS 16               & \textbf{80.78} & 68.68     & 79.98     & \textbf{80.99} & \textbf{80.66} \\ \hline
BS 16 + WD 5e-3     & 71.94     & 66.63     & 80.78     & 77.42     & 79.06 \\ \hline \hline
Base model (theirs) & 79.27     & 67.98     & 80.19     & 79.71     & 79.95 \\ \hline
\end{tabular}
\end{table}

\begin{table}[t]
\centering
\captionsetup{justification=centering}
\caption{Performance metrics comparison between base and optimized model (RAPv1)}
\label{tab:RAP1_bald_optimized}
\vspace{-2mm}
\begin{tabular}{c|c|c|c|c|c}
\hline
\textbf{Metric}                  & \textbf{ma}   & \textbf{Acc}   & \textbf{Prec}   & \textbf{Rec}   & \textbf{F1}   \\ \hline\hline
Base performance                 & 79.27         & 67.98          & \textbf{80.19}  & 79.71        & 79.95         \\ \hline
Performance optimized      & \textbf{80.54}& \textbf{68.49} & 79.89           & \textbf{80.72}& \textbf{80.30} \\
        model       & \textbf{$\pm$0.17}& \textbf{$\pm$0.10}& 0.0765          & \textbf{$\pm$0.13}& \textbf{$\pm$0.08} \\ \hline
\end{tabular}%
\end{table}

\paragraph{Optimization for RAPv2:}
Observing the results for RAPv1, and noting the potential influence of batch size on the optimal settings for other hyperparameters, we extended the previous strategy by incorporating combinations of batch size with each hyperparameter. If any combination yielded improved performance, we proposed to replicate the experiments on RAPv1.

The set of hyperparameters remained unchanged, with the addition of using an EMA model, and the performance improvements were comparable. Notably, modifying only the batch size and weight decay yielded superior results relative to the base model. Table \ref{tab:RAP2_bald_weightDecayBatchSize_optimization} illustrates the joint optimization of these parameters, where a batch size of 16 produced the best overall performance. Although a weight decay of 5e-3 with a batch size of 64 achieved more peak values, the batch size of 16 consistently enhanced all metrics and showed greater stability. The final optimized parameters are listed in Table \ref{tab:RAP2_bald_optimized}.

\begin{table}[t]
\centering
\captionsetup{justification=centering}
\caption{Performance metrics for different weight decays and batch sizes combinations (RAPv2).}
\label{tab:RAP2_bald_weightDecayBatchSize_optimization}
\vspace{-2mm}
\begin{tabular}{c|c|c|c|c|c|c}
\hline
\textbf{Weight Decay} & \textbf{Batch Size} & \textbf{ma}  & \textbf{Acc}         & \textbf{Prec}         & \textbf{Rec}         & \textbf{F1}  \\ 
\hline \hline
5e-3                 & 16         & 71.73      & 65.55                & \textit{79.31}        & 77.43                & 78.36      \\ \hline
5e-3                 & 32         & 75.11      & \textit{67.12}       & \textit{79.57}        & 79.3                 & \textit{79.43} \\ \hline
5e-3                 & 64         & 76.24      & \textbf{\textit{67.84}} & \textbf{\textit{80.01}} & 79.76                & \textbf{\textit{79.89}} \\ \hline
5e-4                 & 16         & \textit{79.06} & \textit{67.18}       & \textit{78.38}        & \textbf{\textit{80.43}} & \textit{79.39} \\ \hline
5e-4                 & 32         & \textit{78.81} & \textit{67.07}       & \textit{78.51}        & 80.07                & \textit{79.28} \\ \hline
5e-4                 & 64         & 77.89      & \textit{66.73}       & \textit{78.7}         & 79.36                & \textit{79.03} \\ \hline
5e-5                 & 16         & \textbf{\textit{79.38}} & \textit{66.84}       & \textit{77.89}        & \textbf{\textit{80.43}} & \textit{79.14} \\ \hline
5e-5                 & 32         & \textit{78.68} & \textit{66.71}       & \textit{78.17}        & 79.95                & \textit{79.05} \\ \hline
5e-5                 & 64         & 78.15      & \textit{66.82}       & \textit{78.58}        & 79.63                & \textit{79.1}  \\ \hline \hline
5e-4 (theirs)                & 64 (theirs) & 78.52      & 66.09                & 77.2                 & 80.23                & 78.68      \\ \hline
\end{tabular}
\end{table}

\begin{table}[t]
\centering
\captionsetup{justification=centering}
\caption{Performance metrics comparison between base and optimized model (RAPv2).}
\label{tab:RAP2_bald_optimized}
\begin{tabular}{c|c|c|c|c|c}
\hline
Performance & ma & Acc & Prec & Rec & F1 \\
\hline
\hline
Base model & 78.52 & 66.09 & 77.2 & 80.23 & 78.68 \\ 
\hline
Optimized & \textbf{79.12} & \textbf{67.17} & \textbf{78.35} & \textbf{80.43} & \textbf{79.38} \\
model & \textbf{± 0.1021} & \textbf{± 0.1037} & \textbf{± 0.0874} & \textbf{± 0.1175} & \textbf{± 0.0869} \\
\hline
\end{tabular}
\end{table}
\vspace{-2mm}

\paragraph{Optimization for RAPZs:}

In the case of RAPzs, after evaluating the trend in the other datasets of the same family, it was decided to focus the optimization process on the batch size. As can be seen in table \ref{tab:RAPzs_bald_BatchSize}, in this case, even smaller values than those being used in the other datasets continue to improve the results, but it was decided to stop at 8 to avoid having too small a batch size that could slow down the training process. The final results are in Table \ref{tab:RAPzs_bald_optimized}.

\begin{table}[t]
\centering
\captionsetup{justification=centering}
\caption{Performance metrics for different batch sizes (RAPzs).}
\label{tab:RAPzs_bald_BatchSize}
\vspace{-2mm}
\begin{tabular}{c|c|c|c|c|c}
\hline
\textbf{Batch size} & \textbf{ma} & \textbf{Acc} & \textbf{Prec} & \textbf{Rec} & \textbf{F1} \\ 
\hline \hline
8               & \textbf{\textit{74.07}} & \textit{65.53} & 78.03 & \textbf{\textit{78.22}} & \textit{78.13} \\ \hline
16              & \textit{73.18}         & \textbf{\textit{65.84}} & \textbf{\textit{79.01}} & \textit{77.69} & \textbf{\textit{78.35}} \\ \hline
32              & \textit{72.2}          & \textit{65.05} & 78.81 & \textit{76.82} & \textit{77.8} \\ \hline
64              & \textit{71.76}         & \textit{64.83} & 78.75 & \textit{76.6} & \textit{77.66} \\ \hline \hline
64 (theirs)     & 71.18                 & 64.54        & 78.85 & 76.02        & 77.41 \\ \hline
\end{tabular}%
\end{table}

\begin{table}[t]
\centering
\caption{Performance metrics comparison between base and optimized model (RAPzs).}
\label{tab:RAPzs_bald_optimized}
\begin{tabular}{c|c|c|c|c|c}
\hline
Performance & ma & Acc & Prec & Rec & F1 \\
\hline
\hline
Base & 71.76 & 64.83 & \textbf{78.75} & 76.6 & 77.66 \\ 
\hline
Optimized & \textbf{74.39} & \textbf{65.59} & 78.19 & \textbf{78.20} & \textbf{78.20} \\
 model & \textbf{± 0.2155} & \textbf{± 0.2281} & ± 0.1459 & \textbf{± 0.2664} & \textbf{± 0.1868} \\
\hline
\end{tabular}
\end{table}

\section{Extended experimental results for Attribute analysis}
\label{sec:extended-experiments-attributes}
Here, we provided further details on the attribute performance (see subsection \ref{annex:attribute_analysis_per_dataset}), the assigned scores using the proposed approach (see subsection \ref{annex:attribute_analysis_per_dataset}), and the identified critical attributes for each dataset (see Subsection \ref{annex:critical_attributes_per_dataset}) and globally (see Subsection \ref{annex:critical_attributes_final_results}). These identified attributes are the best candidates to be improved by using synthetic data.

\subsection{Attribute performance per dataset}
\label{annex:attribute_analysis_per_dataset}
The following Tables provide performance results for Train and Test sets of RAPv1, RAPv2 and RAPzs. We employ all common metrics to measure PAR performance (ma, Acc, Prec, Rec and F1).

\renewcommand{\arraystretch}{0.85}
\begin{table}[!h]
\centering
\caption{RAPv1 training results per attribute}
\label{table:rap1_train_results}
\setlength{\tabcolsep}{3pt}
\begin{tabular}{c|l|c|c|c|c|c|c}
\hline
\textbf{Attribute Number} & \textbf{Attribute Name} & \textbf{ma} & \textbf{Acc} & \textbf{Prec} & \textbf{Rec} & \textbf{F1} & \textbf{Num images} \\
\hline\hline
1  & hs-BaldHead              & 87.28 & 69.47 & 91.00 & 74.59 & 81.98 & 122   \\ \hline
2  & hs-LongHair              & 97.27 & 88.94 & 91.85 & 96.57 & 94.15 & 6348  \\ \hline
3  & hs-BlackHair             & 92.41 & 98.32 & 99.17 & 99.14 & 99.15 & 31437 \\ \hline
4  & hs-Hat                   & 96.41 & 89.07 & 95.58 & 92.90 & 94.22 & 535   \\ \hline
5  & hs-Glasses               & 94.86 & 79.31 & 86.16 & 90.89 & 88.46 & 2459  \\ \hline
6  & hs-Muffler               & 83.43 & 57.49 & 80.13 & 67.04 & 73.01 & 355   \\ \hline
7  & ub-Shirt                 & 96.04 & 87.72 & 92.75 & 94.17 & 93.46 & 7340  \\
\hline
\end{tabular}
\end{table}

\clearpage



\begin{table}[h!]
\centering
\caption{RAPv1 test results per attribute}
\label{table:rap1_test_results}
\setlength{\tabcolsep}{3pt}
\begin{tabular}{c|l|c|c|c|c|c|c}
\hline
\textbf{Attribute Number} & \textbf{Attribute Name} & \textbf{ma} & \textbf{Acc} & \textbf{Prec} & \textbf{Rec} & \textbf{F1} & \textbf{Num images} \\
\hline\hline
1  & hs-BaldHead              & 69.39 & 31.11 & 60.87 & 38.89 & 47.46 & 36 \\ \hline
2  & hs-LongHair              & 93.44 & 78.83 & 86.31 & 90.10 & 88.16 & 1525 \\ \hline
3  & hs-BlackHair             & 83.86 & 96.45 & 98.19 & 98.19 & 98.19 & 7851 \\ \hline
4  & hs-Hat                   & 85.77 & 61.17 & 80.42 & 71.88 & 75.91 & 160 \\ \hline
5  & hs-Glasses               & 88.69 & 64.08 & 77.12 & 79.13 & 78.11 & 575 \\ \hline
6  & hs-Muffler               & 70.28 & 24.09 & 36.67 & 41.25 & 38.82 & 80 \\ \hline
7  & ub-Shirt                 & 88.21 & 68.65 & 80.81 & 82.02 & 81.41 & 1858 \\
\hline
\end{tabular}
\end{table}


\begin{table}[h!]
\centering
\caption{RAPv2 training results per attribute}
\label{table:rap2_train_results}
\setlength{\tabcolsep}{3pt}
\begin{tabular}{c|l|c|c|c|c|c|c}
\hline
\textbf{Attribute Number} & \textbf{Attribute Name} & \textbf{ma} & \textbf{Acc} & \textbf{Prec} & \textbf{Rec} & \textbf{F1} & \textbf{Num images} \\
\hline\hline
1  & hs-BaldHead          & 90.93 & 76.61 & 92.23 & 81.90 & 86.76 & 116 \\
\hline
2  & hs-LongHair          & 96.66 & 87.76 & 91.60 & 95.45 & 93.48 & 4239 \\
\hline
3  & hs-BlackHair         & 90.05 & 98.28 & 99.08 & 99.18 & 99.13 & 20708 \\
\hline
4  & hs-Hat               & 94.39 & 84.29 & 94.27 & 88.85 & 91.48 & 296 \\
\hline
5  & hs-Glasses           & 93.07 & 76.10 & 85.71 & 87.16 & 86.43 & 1426 \\
\hline
6  & ub-Shirt             & 95.47 & 86.70 & 92.23 & 93.53 & 92.88 & 5384 \\
\hline
7  & ub-Sweater           & 82.57 & 52.31 & 67.25 & 70.20 & 68.69 & 2802 \\
\hline
\end{tabular}
\end{table}


\begin{table}[h!]
\centering
\caption{RAPv2 test results per attribute}
\label{table:rap2_test_results}
\setlength{\tabcolsep}{3pt}
\begin{tabular}{c|l|c|c|c|c|c|c}
\hline\hline
\textbf{Attribute Number} & \textbf{Attribute Name} & \textbf{ma} & \textbf{Acc} & \textbf{Prec} & \textbf{Rec} & \textbf{F1} & \textbf{Num images} \\
\hline
1  & hs-BaldHead         & 89.21 & 67.77 & 83.20 & 78.52 & 80.79 & 391 \\
\hline
2  & hs-LongHair         & 96.08 & 85.15 & 89.23 & 94.90 & 91.98 & 13129 \\
\hline
3  & hs-BlackHair        & 87.94 & 96.49 & 98.37 & 98.06 & 98.22 & 63314 \\
\hline
4  & hs-Hat              & 94.47 & 82.71 & 92.06 & 89.06 & 90.54 & 1042 \\
\hline
5  & hs-Glasses          & 93.02 & 71.70 & 79.79 & 87.61 & 83.52 & 4506 \\
\hline
6  & ub-Shirt            & 93.15 & 79.63 & 87.50 & 89.85 & 88.66 & 14701 \\
\hline
7  & ub-Sweater          & 79.83 & 45.25 & 61.66 & 62.96 & 62.30 & 5276 \\
\hline
\end{tabular}
\end{table}


\begin{table}[h!]
\centering
\caption{RAPzs training results per attribute}
\label{table:rapzs_train_results}
\setlength{\tabcolsep}{3pt}
\begin{tabular}{c|l|c|c|c|c|c|c}
\hline
\textbf{Attribute Number} & \textbf{Attribute Name} & \textbf{ma} & \textbf{Acc} & \textbf{Prec} & \textbf{Rec} & \textbf{F1} & \textbf{Num images} \\
\hline\hline
1  & hs-BaldHead             & 81.99 & 50.00 & 69.41 & 64.13 & 66.67 & 92 \\
\hline
2  & hs-LongHair             & 93.71 & 80.18 & 87.54 & 90.51 & 89.00 & 3276 \\
\hline
3  & hs-BlackHair            & 81.91 & 95.48 & 97.60 & 97.78 & 97.69 & 15860 \\
\hline
4  & hs-Hat                  & 88.85 & 66.07 & 81.18 & 78.01 & 79.57 & 282 \\
\hline
5  & hs-Glasses              & 90.88 & 64.35 & 73.41 & 83.91 & 78.31 & 1125 \\
\hline
6  & ub-Shirt                & 88.98 & 68.74 & 79.07 & 84.03 & 81.47 & 3645 \\
\hline
7  & ub-Sweater              & 71.82 & 28.16 & 39.13 & 50.12 & 43.94 & 1303 \\
\hline
\end{tabular}
\end{table}

\begin{table}[h!]
\centering
\caption{RAPzs test results per attribute}
\label{table:rapzs_test_results}
\setlength{\tabcolsep}{3pt}
\begin{tabular}{c|l|c|c|c|c|c|c}
\hline
\textbf{Attribute Number} & \textbf{Attribute Name} & \textbf{ma} & \textbf{Acc} & \textbf{Prec} & \textbf{Rec} & \textbf{F1} & \textbf{Num images} \\
\hline\hline
1  & hs-BaldHead             & 74.93 & 23.08 & 30.00 & 50.00 & 37.50 & 6    \\
\hline
2  & hs-LongHair             & 92.54 & 79.93 & 89.03 & 88.66 & 88.84 & 1217 \\
\hline
3  & hs-BlackHair            & 79.19 & 96.75 & 98.14 & 98.56 & 98.35 & 4709 \\
\hline
4  & hs-Hat                  & 86.19 & 63.10 & 82.81 & 72.60 & 77.37 & 73   \\
\hline
5  & hs-Glasses              & 86.63 & 62.43 & 78.47 & 75.33 & 76.87 & 450  \\
\hline
6  & ub-Shirt                & 86.70 & 67.83 & 83.19 & 78.60 & 80.83 & 1215 \\
\hline
7  & ub-Sweater              & 66.88 & 26.30 & 38.75 & 45.02 & 41.65 & 673  \\
\hline
\end{tabular}
\end{table}

\clearpage
\newpage  

\subsection{Attribute scores per dataset}
\label{annex:attribute_analysis_per_dataset}

\renewcommand{\arraystretch}{0.90}
\begin{table}[h!]
\centering
\caption{Attribute Scores for RAPv1}
\label{table:attribute_analysis_results_rap1}
\setlength{\tabcolsep}{3pt}
\begin{tabular}{c|l|c|c|c|c}
\hline
\textbf{Attribute Number} & \textbf{Attribute Name} & \textbf{Low Training Images} & \textbf{Test performance} & \textbf{Drop in Performance} & \textbf{Score} \\
\hline\hline
1  & hs-BaldHead           & True  & Low    & Big    & 0 \\
\hline
2  & hs-LongHair           & False & High   & Small  & 6 \\
\hline
3  & hs-BlackHair          & False & High   & Small  & 6 \\
\hline
4  & hs-Hat                & True  & Medium & Medium & 2 \\
\hline
5  & hs-Glasses            & False & Medium & Small  & 5 \\
\hline
6  & hs-Muffler            & True  & Low    & Big    & 0 \\
\hline
7  & ub-Shirt              & False & High   & Small  & 6 \\
\hline
\end{tabular}
\end{table}

\clearpage

\begin{table}[h!]
\centering
\caption{Attribute Scores for RAPv2}
\label{table:attribute_analysis_results_rap2}
\setlength{\tabcolsep}{3pt}
\begin{tabular}{c|l|c|c|c|c}
\hline
\textbf{Attribute Number} & \textbf{Attribute Name} & \textbf{Low Training Images} & \textbf{Test performance} & \textbf{Drop in Performance} & \textbf{Score} \\
\hline\hline
1  & hs-BaldHead            & True  & Medium & Small  & 3  \\ \hline
2  & hs-LongHair            & False & High   & Small  & 6  \\ \hline
3  & hs-BlackHair           & False & High   & Small  & 6  \\ \hline
4  & hs-Hat                 & True  & Medium & Small  & 3  \\ \hline
5  & hs-Glasses             & False & Medium & Small  & 5  \\ \hline
6  & ub-Shirt               & False & High   & Small  & 6  \\ \hline
7  & ub-Sweater             & False & Low    & Medium & 3  \\
\hline
\end{tabular}
\end{table}

\begin{table}[h!]
\centering
\caption{Attribute Scores for RAPzs}
\label{table:attribute_analysis_results_rapzs}
\setlength{\tabcolsep}{3pt}
\begin{tabular}{c|l|c|c|c|c}
\hline
\textbf{Attribute Number} & \textbf{Attribute Name} & \textbf{Low Training Images} & \textbf{Test performance} & \textbf{Drop in Performance} & \textbf{Score} \\
\hline\hline
1  & hs-BaldHead            & True  & Low    & Big    & 0  \\\hline
2  & hs-LongHair            & False & High   & Small  & 6  \\\hline
3  & hs-BlackHair           & False & High   & Small  & 6  \\\hline
4  & hs-Hat                 & True  & Medium & Small  & 3  \\\hline
5  & hs-Glasses             & False & Medium & Small  & 5  \\\hline
6  & ub-Shirt               & False & High   & Small  & 6  \\\hline
7  & ub-Sweater             & False & Low    & Medium & 3  \\
\hline
\end{tabular}
\end{table}

\clearpage
\newpage

\twocolumn

\subsection{Critical attributes per dataset}
\label{annex:critical_attributes_per_dataset}

\begin{table}[h!]
\centering
\captionsetup{justification=centering}
\caption{Score per attribute RAPv1}
\vspace{-2mm}
\resizebox{\columnwidth}{!}{%
\begin{tabular}{c|c|l}
\hline
\textbf{Score} & \textbf{Attribute Number} & \textbf{Attribute Name} \\ \hline \hline
0                  & 1  & hs-BaldHead \\ \cline{2-3}
                   & 6  & hs-Muffler \\ \cline{2-3}
                   & 47 & action-Holding \\ \cline{2-3}
                   & 50 & action-CarrybyArm \\ \hline
1                  & 14 & ub-Tight \\ \cline{2-3}
                   & 32 & attach-PaperBag \\ \hline
2                   & 4  & hs-Hat \\
\hline
\end{tabular}%
}
\label{table:score_attributes_rap1}
\end{table}

\begin{table}[h!]
\centering
\captionsetup{justification=centering}
\caption{Score per attribute RAPv2}
\vspace{-2mm}
\resizebox{\columnwidth}{!}{%
\begin{tabular}{c|c|l}
\hline
\textbf{Score} & \textbf{Attribute Number} & \textbf{Attribute Name} \\ \hline \hline
1                  & 49 & action-Holding \\ \hline
2                  & 15 & ub-Others \\ \cline{2-3}
                   & 17 & lb-Skirt \\ \cline{2-3}
                   & 18 & lb-ShortSkirt \\ \cline{2-3}
                   & 19 & lb-Dress \\ \cline{2-3}
                   & 30 & attachment-HandBag \\ \cline{2-3}
                   & 32 & attachment-PlasticBag \\
\hline
\end{tabular}%
}
\label{table:score_attributes_rap2}
\end{table}

\begin{table}[h!]
\centering
\captionsetup{justification=centering}
\caption{Score per attribute RAPzs}
\vspace{-2mm}
\resizebox{\columnwidth}{!}{%
\begin{tabular}{c|c|l}
\hline
\textbf{Score} & \textbf{Attribute Number} & \textbf{Attribute Name} \\ \hline \hline
0                  & 1  & hs-BaldHead \\ \cline{2-3}
                   & 14 & ub-ShortSleeve \\ \cline{2-3}
                   & 25 & shoes-Cloth \\ \cline{2-3}
                   & 48 & action-Gathering \\ \cline{2-3}
                   & 49 & action-Holding \\ \cline{2-3}
                   & 51 & action-Pulling \\ \hline
1                  & 18 & lb-ShortSkirt \\
\hline
\end{tabular}%
}
\label{table:score_attributes_rapzs}
\end{table}

\clearpage

\onecolumn

\subsection{Critical attributes final results}
\label{annex:critical_attributes_final_results}

\begin{table}[!htbp]
\centering
\footnotesize
\caption{Top 20 Worse Attributes per Dataset}
\label{table:top_worse_attributes_per_dataset}
\setlength{\tabcolsep}{3pt}
\begin{tabular}{c|c|l|c|l|c|l}
\hline
\textbf{Lower} & 
\shortstack{\textbf{RAP1}\\\textbf{Attribute Number}} & 
\shortstack{\textbf{RAP1}\\\textbf{Attribute Name}} & 
\shortstack{\textbf{RAP2}\\\textbf{Attribute Number}} & 
\shortstack{\textbf{RAP2}\\\textbf{Attribute Name}} & 
\shortstack{\textbf{RAPZS}\\\textbf{Attribute Number}} & 
\shortstack{\textbf{RAPZS}\\\textbf{Attribute Name}} \\
\hline\hline
1  & 1  & hs-BaldHead          & 49 & action-Holding         & 1  & hs-BaldHead \\
\hline
2  & 6  & hs-Muffler           & 15 & ub-Others              & 14 & ub-ShortSleeve \\
\hline
3  & 47 & action-Holding       & 17 & lb-Skirt               & 25 & shoes-Cloth \\
\hline
4  & 50 & action-CarrybyArm    & 18 & lb-ShortSkirt          & 48 & action-Gathering \\
\hline
5  & 14 & ub-Tight             & 19 & lb-Dress               & 49 & action-Holding \\
\hline
6  & 32 & attach-PaperBag      & 30 & attachment-HandBag     & 51 & action-Pulling \\
\hline
7  & 4  & hs-Hat               & 32 & attachment-PlasticBag  & 18 & lb-ShortSkirt \\
\hline
\end{tabular}
\end{table}

\clearpage

\begin{table}[h!]
\centering
\caption{Number of datasets where each attribute is on the top-worse attributes}
\label{table:attribute_selection_final_results}
\setlength{\tabcolsep}{3pt}
\begin{tabular}{c|l|c}
\hline
\textbf{Attribute Number} & \textbf{Attribute Name} & \textbf{Times as Top-Worse Attribute} \\
\hline\hline
49 & action-Holding & 3 \\
\hline
51 & action-Pulling & 3 \\
\hline
1  & hs-BaldHead   & 3 \\
\hline
18 & lb-ShortSkirt & 3 \\
\hline
36 & AgeLess16     & 3 \\
\hline
4  & hs-Hat        & 3 \\
\hline
25 & shoes-Cloth   & 3 \\
\hline
\end{tabular}
\end{table}

\section{Extended experimental results for Synthetic Data Augmentation}
\label{sec:extended-experiments-dataaugmentation}

\subsection{Parameter Selection for Generation}
To optimize augmentation, we selected the weakest attribute \textit{hs-BaldHead} identified by our approach, and we evaluated three key parameters: image noise level, augmented weight, and the number of images added to the dataset, represented as the percentage of number of images added relative to the number of images the attribute has in the training set. For example, \textit{hs-BaldHead} in RAPv1 has $122$ images in training; an augmentation percentage of $50\%$ represents adding $66$ synthetic images to the training set with this attribute.

The augmented weight is designed to influence the dataset's overall results, as its value is only parameterizable for attributes labeled as $3$ (the indirectly augmented ones). For this reason, finding an optimal value for it has been reserved for later.

With this in mind, the first step was to analyze the relationship between the level of noise in the image and the percentage of augmentation applied to the attribute, keeping the augmented weight fixed at $0.5$. Based on the results yielding the best performance, further experiments were proposed to observe the effect of varying the augmented weight. For this purpose, three levels of noise were configured, modifying how much blending of the generated noise occurs in the image (at $25\%$, $50\%$, or $75\%$). Initially, augmentation levels of $50\%$, $100\%$, $200\%$, and $300\%$ were configured. However, after observing the initial results, additional values between $400\%$ and $700\%$ were tested.

The training processes were conducted to ensure that the only differences between the various parameter configurations were those directly influencing the parameters themselves. Specifically, noise levels were added to the same base image, meaning that image 1 of \textit{small\_noise} and image 1 of \textit{big\_noise} are identical except for the noise level. Similarly, augmentation was applied using overlapping sets, such that $100\%$ augmentation includes as first half the $50\%$ set, $200\%$ includes the $100\%$ set, and so forth. All results computed are available in Tables \ref{table:hs_baldhead_calculations}, \ref{table:rapv2_baldhead} and \ref{table:rapzs_baldhead}.

\begin{table}[h!]
\centering
\caption{Performance results for data augmentation of \textit{hs-BaldHead} in RAPv1}
\label{table:hs_baldhead_calculations}
\setlength{\tabcolsep}{3pt}
\begin{tabular}{c|c|c|c|c|c|c|c|c|c|c|c|c}
\hline
\textbf{\% Augmentation} & \textbf{Noise} & \textbf{Aug. Weight} & \multicolumn{5}{c|}{\textbf{BaldHead}} & \multicolumn{5}{c}{\textbf{All Attributes}} \\
\cline{4-13}
 & & & \textbf{ma} & \textbf{Acc} & \textbf{Prec} & \textbf{Rec} & \textbf{F1} & \textbf{ma} & \textbf{Acc} & \textbf{Prec} & \textbf{Rec} & \textbf{F1} \\
\hline\hline
50  & Big    & 0.5 & 73.56 & 37.78 & 65.38 & 47.22 & 54.84 & 80.50 & 68.43 & 80.03 & 80.54 & 80.28 \\ \hline
100 & Big    & 0.5 & 81.87 & 46.94 & 63.89 & 63.89 & 63.89 & 81.15 & 68.39 & 79.76 & 80.75 & 80.25 \\ \hline
200 & Big    & 0.5 & 77.71 & 42.55 & 64.52 & 55.56 & 59.70 & 80.51 & 68.74 & 80.05 & 80.88 & 80.46 \\ \hline
300 & Big    & 0.5 & 77.71 & 41.67 & 62.50 & 55.56 & 58.82 & 80.48 & 68.27 & 79.68 & 80.69 & 80.18 \\ \hline
400 & Big    & 0.5 & 77.70 & 40.82 & 60.61 & 55.56 & 57.97 & 80.91 & 68.47 & 79.64 & 81.02 & 80.33 \\ \hline
500 & Big    & 0.5 & 73.54 & 36.17 & 60.71 & 47.22 & 53.12 & 80.41 & 68.27 & 79.40 & 80.91 & 80.14 \\ \hline
600 & Big    & 0.5 & 77.72 & 43.48 & 66.67 & 55.56 & 60.61 & 80.95 & 68.65 & 79.86 & 81.07 & 80.46 \\ \hline
700 & Big    & 0.5 & 76.31 & 38.78 & 59.38 & 52.78 & 55.88 & 80.94 & 68.54 & 79.54 & 81.25 & 80.39 \\ \hline
\hline
50  & Medium & 0.5 & 76.31 & 38.78 & 59.38 & 52.78 & 55.88 & 80.56 & 68.48 & 79.92 & 80.75 & 80.34 \\\hline
100 & Medium & 0.5 & 77.72 & 43.48 & 66.67 & 55.56 & 60.61 & 80.62 & 68.60 & 80.03 & 80.72 & 80.37 \\\hline
200 & Medium & 0.5 & 77.72 & 44.44 & 68.97 & 55.56 & 61.54 & 80.24 & 68.54 & 79.98 & 80.72 & 80.35 \\\hline
300 & Medium & 0.5 & 77.71 & 42.55 & 64.52 & 55.56 & 59.70 & 80.84 & 68.88 & 79.96 & 81.29 & 80.62 \\\hline
400 & Medium & 0.5 & 77.72 & 44.44 & 68.97 & 55.56 & 61.54 & 80.83 & 68.34 & 79.78 & 80.65 & 80.21 \\\hline
500 & Medium & 0.5 & 77.74 & 46.51 & 74.07 & 55.56 & 63.49 & 80.97 & 68.65 & 79.56 & 81.38 & 80.46 \\\hline
600 & Medium & 0.5 & 74.93 & 38.30 & 62.07 & 50.00 & 55.38 & 81.09 & 68.92 & 79.97 & 81.32 & 80.64 \\\hline
700 & Medium & 0.5 & 74.91 & 35.29 & 54.55 & 50.00 & 52.17 & 80.77 & 68.71 & 79.62 & 81.37 & 80.49 \\\hline
\hline
50  & Small  & 0.5 & 74.94 & 39.13 & 64.29 & 50.00 & 56.25 & 80.84 & 68.68 & 79.99 & 80.95 & 80.47 \\\hline
100 & Small  & 0.5 & 80.49 & 46.81 & 66.67 & 61.11 & 63.77 & 80.87 & 68.46 & 79.94 & 80.67 & 80.30 \\\hline
200 & Small  & 0.5 & 77.71 & 41.67 & 62.50 & 55.56 & 58.82 & 80.38 & 68.60 & 79.86 & 80.90 & 80.37 \\\hline
300 & Small  & 0.5 & 79.11 & 46.67 & 70.00 & 58.33 & 63.64 & 80.41 & 68.44 & 79.85 & 80.70 & 80.27 \\\hline
400 & Small  & 0.5 & 74.92 & 36.00 & 56.25 & 50.00 & 52.94 & 80.81 & 68.44 & 79.68 & 80.89 & 80.28 \\\hline
500 & Small  & 0.5 & 73.58 & 41.46 & 77.27 & 47.22 & 58.62 & 80.32 & 68.26 & 79.31 & 80.97 & 80.13 \\\hline
600 & Small  & 0.5 & 73.56 & 37.78 & 65.38 & 47.22 & 54.84 & 80.77 & 68.74 & 79.91 & 81.11 & 80.51 \\\hline
700 & Small  & 0.5 & 74.95 & 40.91 & 69.23 & 50.00 & 58.06 & 80.61 & 68.48 & 79.56 & 81.11 & 80.32 \\\hline
\end{tabular}
\end{table}

\begin{table}[h!]
\centering
\caption{Performance results for data augmentation of \textit{hs-BaldHead} in RAPv2}
\label{table:rapv2_baldhead}
\setlength{\tabcolsep}{3pt}
\begin{tabular}{c|c|c|c|c|c|c|c|c|c|c|c|c}
\hline
\textbf{\% Augmentation} & \textbf{Noise} & \textbf{Aug. Weight} & \multicolumn{5}{c|}{\textbf{BaldHead}} & \multicolumn{5}{c}{\textbf{All Attributes}} \\
\cline{4-13}
 &  &  & \textbf{ma} & \textbf{Acc} & \textbf{Prec} & \textbf{Rec} & \textbf{F1} & \textbf{ma} & \textbf{Acc} & \textbf{Prec} & \textbf{Rec} & \textbf{F1} \\
\hline\hline
50  & Medium & 0.5 & 80.62 & 46.905 & 66.68  & 61.41 & 63.86  & 78.88  & 67.265 & 78.59  & 80.33 & 79.45 \\ \hline
100 & Medium & 0.5 & 81.46 & 51.79  & 74.36  & 63.04 & 68.24  & 78.96  & 67.16  & 78.16  & 80.64 & 79.38 \\ \hline
200 & Medium & 0.5 & 83.33 & 49.395 & 65.425 & 66.845 & 66.13  & 79.195 & 67.375 & 78.48  & 80.65 & 79.55 \\ \hline
300 & Medium & 0.5 & 82.53 & 50.42  & 68.97  & 65.22  & 67.04  & 79.18  & 67.30  & 78.06  & 80.98 & 79.49 \\ \hline
400 & Medium & 0.5 & 79.55 & 47.215 & 69.935 & 59.24  & 64.135 & 79.135 & 67.67  & 78.21  & 81.42 & 79.78 \\ \hline
500 & Medium & 0.5 & 83.08 & 51.69  & 70.11  & 66.30  & 68.16  & 79.50  & 67.58  & 78.02  & 81.49 & 79.72 \\ \hline
\end{tabular}
\end{table}

\begin{table}[h!]
\centering
\caption{Performance results for data augmentation of \textit{hs-BaldHead} in RAPzs}
\label{table:rapzs_baldhead}
\setlength{\tabcolsep}{3pt}
\begin{tabular}{c|c|c|c|c|c|c|c|c|c|c|c|c}
\hline
\textbf{\% Augmentation} & \textbf{Noise} & \textbf{Aug. Weight} & \multicolumn{5}{c|}{\textbf{BaldHead}} & \multicolumn{5}{c}{\textbf{All Attributes}} \\
\cline{4-13}
 &  &  & \textbf{ma} & \textbf{Acc} & \textbf{Prec} & \textbf{Rec} & \textbf{F1} & \textbf{ma} & \textbf{Acc} & \textbf{Prec} & \textbf{Rec} & \textbf{F1} \\
\hline\hline
50  & Medium & 0.5 & 66.61 & 16.67 & 25.00 & 33.33 & 28.57 & 73.42 & 65.82 & 78.58 & 78.09 & 78.34 \\ \hline
100 & Medium & 0.5 & 74.93 & 23.08 & 30.00 & 50.00 & 37.50 & 74.16 & 65.37 & 77.92 & 78.20 & 78.06 \\ \hline
200 & Medium & 0.5 & 66.60 & 15.38 & 22.22 & 33.33 & 26.67 & 73.67 & 65.06 & 77.68 & 77.91 & 77.79 \\ \hline
300 & Medium & 0.5 & 66.64 & 22.22 & 40.00 & 33.33 & 36.36 & 73.53 & 65.45 & 77.78 & 78.41 & 78.09 \\ \hline
400 & Medium & 0.5 & 66.64 & 22.22 & 40.00 & 33.33 & 36.36 & 73.56 & 65.72 & 78.22 & 78.32 & 78.27 \\ \hline
500 & Medium & 0.5 & 66.65 & 25.00 & 50.00 & 33.33 & 40.00 & 73.49 & 65.39 & 77.84 & 78.28 & 78.06 \\ \hline
\end{tabular}
\end{table}

\begin{figure}[htb]
    \centering
    \includegraphics[width=0.75\columnwidth]{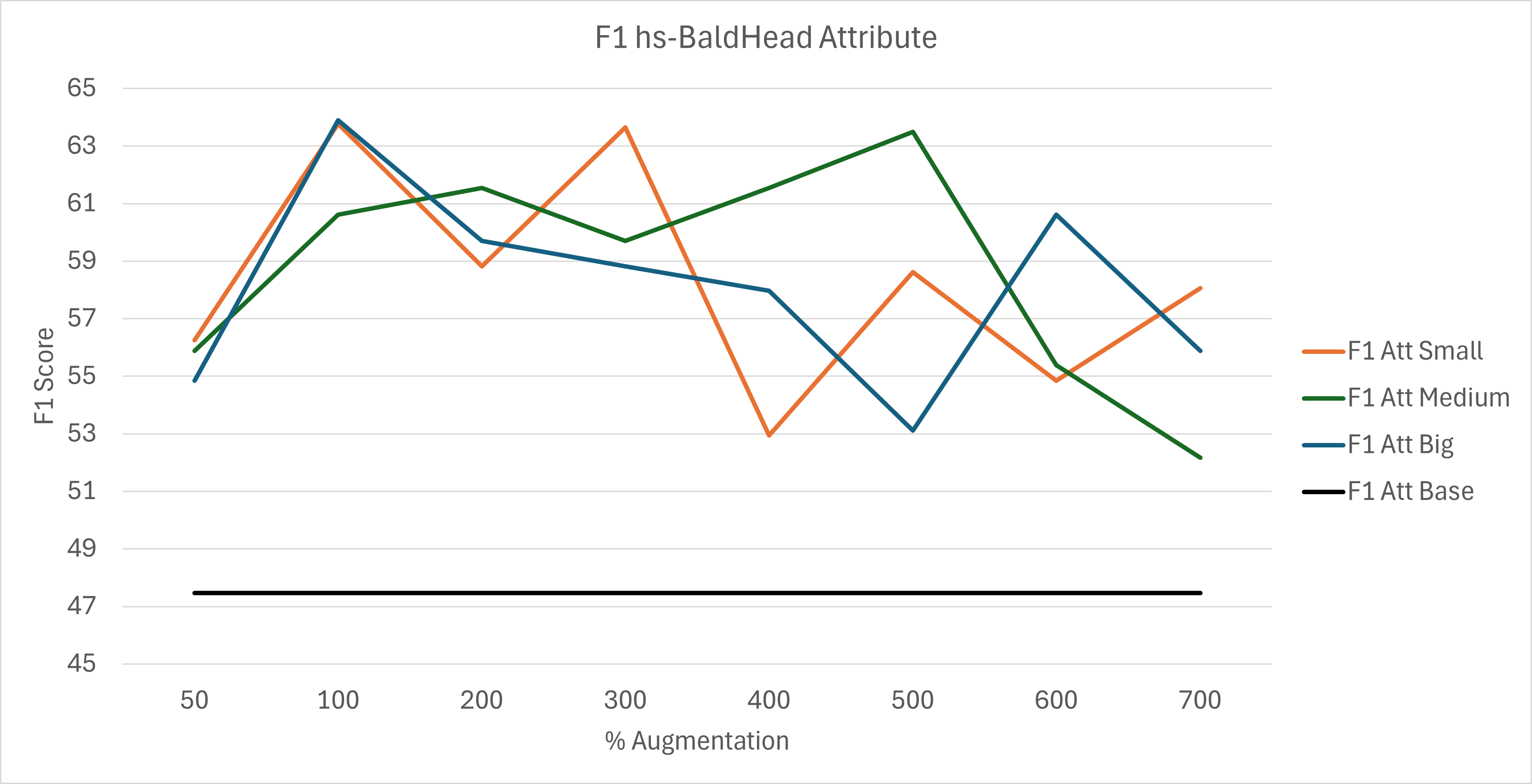}
    \caption{Results for hs-BaldHead depending on the noise level for each augmentation percentage on the attribute}
    \label{fig:baldHead_Attribute_noise}
\end{figure}

The first parameter to set was the augmentation percentage. Figure \ref{fig:baldHead_Attribute_noise} shows the results obtained on hs-BaldHead on RAPv1 for all different levels of noise on the previously mentioned percentages. Based on these observations, even if improvements are achieved in all cases, with some being almost $20$ points better in the selected metric, it was decided to restrict the augmentation percentage to $500\%$, since higher values do not yield improvements and significantly reduce the number of images to generate and the generation time required.

Regarding each noise levels, small and big noise exhibit less stable behavior, with their minimum and maximum values falling within similar ranges to the other augmentations but without displaying clear trends. On the other hand, medium noise appears to show a trend of improvement as more images are added, which is expected, until reaching $500\%$ augmentation, at which point a decline begins.

As shown in Table \ref{tab:all_bald_noise}, for the specific attribute evaluated, the medium noise level yields the highest average performance. This finding demonstrates that the introduction of noise is beneficial, indicating that a minimal amount of noise is insufficient, while simultaneously confirming that an excessive noise level degrades performance.

\begin{table}[t]
\centering
\caption{Attribute-level F1-score comparison of noise levels for \textit{hs-BaldHead} attribute on the RAP datasets. An augmentation weight of 0.5 was employed.}
\vspace{-2mm}
\label{tab:all_bald_noise}
\begin{tabular}{l|c|c|c}
\hline
\textbf{Noise level} & \textbf{RAPv1} & \textbf{RAPv2} & \textbf{RAPzs}\\
\hline
\hline
Big noise (75\%) & 53.12 & 66.67 & 23.53\\ \hline
Medium noise (50\%) & \textbf{63.49} & \textbf{68.16} & \textbf{40.00}\\ \hline
Small noise (25\%)& 58.62 & 66.29 & \textbf{40.00}\\ \hline
\hline
Optimized baseline results & 47.46 & 66.67 & 37.50 \\
\hline\end{tabular}
\vspace{-2mm}
\end{table}

For these reasons, it was deceided to fix the maximum augmentation percentage in 500\%, and the level of noise in the medium one, and with these configuration, the next step was to search for an optimal augmented weight. Various augmented weights between $0$ and $1$ were tested, resulting in the outcomes shown in Table \ref{tab:all_bald_augmentedWeight}.

\begin{table}[t]
\centering
\caption{Attribute-level F1-score comparison of augmentation weights for \textit{hs-BaldHead} attribute on the RAP datasets. A medium noise level was employed.}
\vspace{-2mm}
\label{tab:all_bald_augmentedWeight}
\begin{tabular}{c|c|c|c}
\hline
\textbf{Augmentation Weight} & \textbf{RAPv1}  & \textbf{RAPv2}  & \textbf{RAPzs}\\
\hline
\hline
0.25 & 61.29 & 65.17 & 14.29 \\ \hline
0.5 & \textbf{63.49} & \textbf{68.16} & \textbf{40.00} \\ \hline
0.75 & 53.33 & 64.77 & \textbf{40.00} \\ \hline
 1 & 54.24 & 65.90 & 18.18 \\ \hline
\hline
Optimized baseline results & 47.46 & 66.67 & 37.50 \\
\hline
\end{tabular}
\vspace{-2mm}
\end{table}

As can be observed, the best result is obtained with the default value previously tested. This confirms that the decision to treat attributes labeled as $3$ with different weights is well-founded, as scores degrade when these attributes are given either full or zero weight. Also, it can a be noted how indeed the other attributes are somehow noisy or not fully representative, and therefore 1 is not the best value for this parameter.

In any case, these final results conclude the selection of general hyperparameters for obtaining results on the other datasets. The selected configuration includes medium noise, an augmented weight of $0.5$, and augmentation percentages of $50\%$, $100\%$, $200\%$, $300\%$, $400\%$ and $500\%$.

\vspace{-2mm}
\paragraph{On hs-BaldHead results:}

\begin{table*}[t]
\centering
\caption{Comparison of F1 on the attribute and F1 on the Dataset for the hs-BaldHead attribute across RAPv1, RAPv2, and RAPzs datasets for different data augmentation (DA) percentages.}
\label{tab:all_bald_results}
\setlength{\tabcolsep}{3pt}
\begin{tabular}{c|c|c|c|c|c|c}
\hline
\textbf{DA} & \multicolumn{2}{|c}{\textbf{RAPv1}} & \multicolumn{2}{|c}{\textbf{RAPv2}} & \multicolumn{2}{|c}{\textbf{RAPzs}} \\ \cline{2-7}
      (\%)           & \textbf{F1 Attribute} & \textbf{F1 Dataset} & \textbf{F1 Attribute} & \textbf{F1 Dataset} & \textbf{F1 Attribute} & \textbf{F1 Dataset} \\ \hline \hline
50   & 55.88                & 80.34               & 63.86                & 79.45               & 28.57                & \textbf{78.34}      \\ \hline
100  & 60.61                & 80.37               & \textbf{68.24}       & 79.38               & 37.50                & 78.06               \\ \hline
200  & 61.54                & 80.35               & 66.13                & 79.55               & 26.67                & 77.79               \\ \hline
300  & 59.70                & \textbf{80.62}      & 67.04                & 79.49               & 36.36                & 78.09               \\ \hline
400  & 61.54                & 80.21               & 64.14                & \textbf{79.78}      & 36.36                & 78.27               \\ \hline
500  & \textbf{63.49}       & 80.46               & 68.16                & 79.72               & \textbf{40.00}       & 78.06               \\ \hline\hline
Optimized baseline results & 47.46          & 80.30               & 66.67                & 79.38               & 37.50                & 78.20               \\ \hline
\end{tabular}
\end{table*}

Overall, Table \ref{tab:all_bald_results} presents the final results on the attribute, revealing that incorporating augmentation consistently enhances performance across all datasets. For both RAPv1 and RAPv2, similar trends are observed, with notable performance gains achieved at 500\% augmentation, although for RAPv2 a slight edge is seen at 100\% augmentation.

In contrast, the zero-shot setting exhibits much less pronounced improvements. While the optimal augmentation percentage remains consistent with the previous cases, the overall performance boost relative to the baseline is minimal, particularly given the low starting performance.

\begin{figure}[b]
    \centering
    \includegraphics[width=0.85\columnwidth]{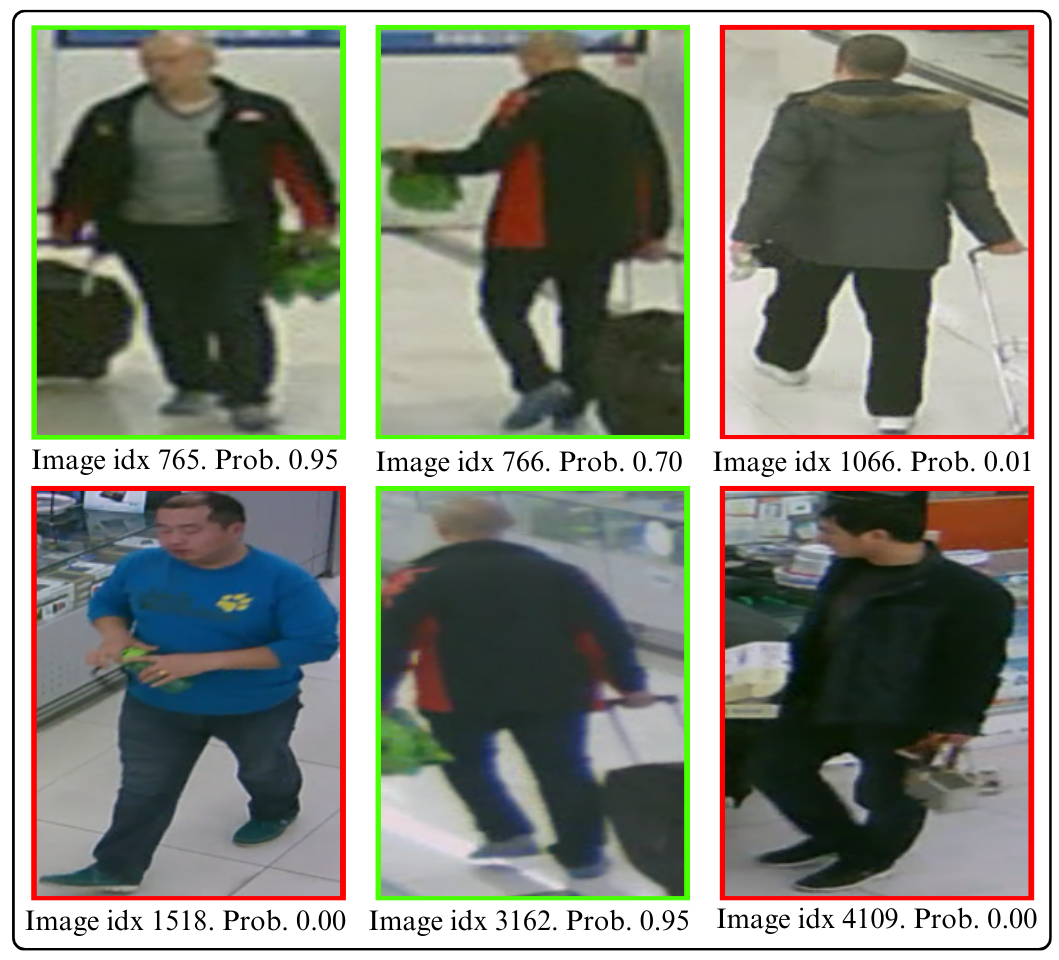}
    \caption{Groundtruth and predicted probabilities for hs-BaldHead in RAPzs, highlighted in green or red based on classification.}
    \label{fig:zs_bald_gt}
\end{figure}

A likely contributing factor is that, as shown in Table \ref{tab:dataset_summary_baldhead}, the \textit{hs-BaldHead} class comprises only 6 images in the testing set. To further investigate the model’s failure modes and assess their representativeness, we examined the ground truth annotations and analyzed the predicted probabilities, with the corresponding findings presented in Figure \ref{fig:zs_bald_gt}.

\begin{table}[t]
\centering
\captionsetup{justification=centering}
\caption{Number of hs-BaldHead train and test images for RAP datasets.}
\label{tab:all_bald_nImages}
\vspace{-2mm}
\begin{tabular}{c|c|c}
\hline
\textbf{Dataset} & \textbf{\# hs-BaldHead Train Images} & \textbf{\# hs-BaldHead Test Images} \\ 
\hline \hline
RAPv1 & 122 & 36 \\ \hline
RAPv2 & 391 & 92 \\ \hline
RAPzs & 116 & 6 \\ \hline
\end{tabular}
\label{tab:dataset_summary_baldhead}
\end{table}

As observed, the model correctly predicts half of the ground truth images, which aligns with the reported metrics. However, a closer examination of the ground truth reveals that, despite the testing set comprising only 6 images, the dataset is noisy; notably, 3 of these images depict individuals with hair.
For instance, image 1518 (bottom left) might be considered bald since, despite the presence of hair, the scalp is visible at the top of the head. In contrast, the two images in the rightmost column clearly show pedestrians with full heads of hair, one of which features untrimmed, short hair.
Thus, the inability of the model to achieve better results on RAPzs is justified. The combination of a very small test set and significant noise in the annotations contributes to the observed discrepancies between the predictions and the expected ground truth.

\vspace{-2mm}
\paragraph{Results on lb-ShortSkirt:}

The attribute \textit{lb-ShortSkirt} was also one of the most challenging for the model to learn. The entire procedure for augmenting this attribute and attempting to improve its results has followed the same general approach as for the final results in \textit{hs-BaldHead}.
However, the first difference arises during the image generation process. As shown in Table \ref{tab:all_shortskirt_nImages}, the number of images composing the augmentation is significantly different. While \textit{hs-BaldHead} had a maximum of $391$ images in training ($1955$ images in total to generate for the maximum augmentation), \textit{lb-ShortSkirt} has more than four times that number. This increases the demand for generated images and highlights a different challenge in learning this attribute, as the base number of training images should be able to give some better base scores.

\begin{table}[t]
\centering
\captionsetup{justification=centering}
\caption{Number of lb-ShortSkirt train and test images for RAP datasets.}
\label{tab:all_shortskirt_nImages}
\vspace{-2mm}
\begin{tabular}{c|c|c}
\hline
\textbf{Dataset} & \textbf{\# lb-ShortSkirt Train Images} & \textbf{\# lb-ShortSkirt Test Images} \\ 
\hline \hline
RAPv1          & 912                                  & 252                                  \\ \hline
RAPv2          & 1390                                 & 354                                  \\ \hline
RAPzs          & 643                                  & 144                                  \\ \hline
\end{tabular}
\end{table}

It is worth noting that another key difference in the generation process lies in the \_\_styles\_\_ wildcards. Several values had to be modified, as the styles, while suitable in all cases for bald individuals, sometimes referenced garments for the lower part of the body. Since \textit{lb-ShortSkirt} belongs to the "lower body" category, these styles created conflicts. For instance, when instructing the model to generate pedestrians with short skirts and pants, it produced examples not aligned with expectations. 

After adjusting the styles, a complete generation was conducted finishing the prompt with " wearing a short skirt" for the attribute. However, upon manually verifying the presence of the attribute in the images for some of the augmentated subsets, it was observed that images generated by the model consistently depicted bare legs, whereas those in the dataset were almost always covered, typically by leggings. Consequently, the prompt was modified to "wearing a short skirt with tights or leggings." This updated prompt not only references the primary attribute but also aligns the generation of the rest of the pedestrian's appearance with the examples in the dataset. 

The results obtained from this augmentation for each dataset are available in Tables \ref{table:shortskirt_calculations}, \ref{table:rapv2_shortskirt}, \ref{tab:rapzs_shortskirt}. Table \ref{tab:all_shortskirt_results} summarizes these results. As can be observed, the results obtained for the attribute itself are worse than those for \textit{hs-BaldHead}. In general, since this attribute starts with higher baseline scores, it is more challenging to improve. However, the scores achieved for the attribute are actually lower than in the base case. This suggests that the generated images fail to capture the patterns that the network needs to learn in order to correctly classify the test images.

\begin{table}[h!]
\centering
\caption{Performance results for data augmentation of \textit{lb-shortSkirt} in RAPv1}
\label{table:shortskirt_calculations}
\setlength{\tabcolsep}{3pt}
\begin{tabular}{c|c|c|c|c|c|c|c|c|c|c|c|c}
\hline
\textbf{\% Augmentation} & \textbf{Noise} & \textbf{Aug. Weight} & \multicolumn{5}{c|}{\textbf{ShortSkirt}} & \multicolumn{5}{c}{\textbf{All Attributes}} \\
\cline{4-13}
 &  &  & \textbf{ma} & \textbf{Acc} & \textbf{Prec} & \textbf{Rec} & \textbf{F1} & \textbf{ma} & \textbf{Acc} & \textbf{Prec} & \textbf{Rec} & \textbf{F1} \\
\hline\hline
50  & Medium & 0.5 & 81.09 & 48.77 & 68.24 & 63.10 & 65.57 & 80.52 & 68.71 & 79.94 & 81.07 & 80.50 \\ \hline
100 & Medium & 0.5 & 81.59 & 52.10 & 73.85 & 63.89 & 68.51 & 81.17 & 68.71 & 79.60 & 81.46 & 80.52 \\ \hline
200 & Medium & 0.5 & 80.93 & 49.38 & 69.91 & 62.70 & 66.11 & 81.04 & 68.67 & 79.21 & 81.84 & 80.51 \\ \hline
300 & Medium & 0.5 & 79.41 & 48.70 & 72.82 & 59.52 & 65.50 & 80.79 & 68.14 & 78.61 & 81.63 & 80.09 \\ \hline
400 & Medium & 0.5 & 80.62 & 51.15 & 74.64 & 61.90 & 67.68 & 81.04 & 68.81 & 79.00 & 82.33 & 80.63 \\ \hline
500 & Medium & 0.5 & 77.47 & 46.51 & 74.07 & 55.56 & 63.49 & 81.01 & 68.64 & 78.29 & 82.90 & 80.53 \\ \hline
\end{tabular}
\vspace{-5mm}
\end{table}

\begin{table}[h!]
\centering
\caption{Performance results for data augmentation of \textit{lb-shortSkirt} in RAPv2}
\label{table:rapv2_shortskirt}
\setlength{\tabcolsep}{3pt}
\begin{tabular}{c|c|c|c|c|c|c|c|c|c|c|c|c}
\hline
\textbf{\% Augmentation} & \textbf{Noise} & \textbf{Aug. Weight} & \multicolumn{5}{c|}{\textbf{ShortSkirt}} & \multicolumn{5}{c}{\textbf{All Attributes}} \\
\cline{4-13}
 &  &  & \textbf{ma} & \textbf{Acc} & \textbf{Prec} & \textbf{Rec} & \textbf{F1} & \textbf{ma} & \textbf{Acc} & \textbf{Prec} & \textbf{Rec} & \textbf{F1} \\
\hline\hline
50  & Medium & 0.5 & 79.17 & 44.56 & 64.51 & 59.04 & 61.65 & 79.42 & 67.17 & 78.00 & 80.83 & 79.39 \\ \hline
100 & Medium & 0.5 & 77.89 & 42.19 & 62.50 & 56.50 & 59.35 & 79.24 & 67.29 & 78.29 & 80.70 & 79.48 \\ \hline
200 & Medium & 0.5 & 78.70 & 42.39 & 60.95 & 58.19 & 59.54 & 79.19 & 67.68 & 78.07 & 81.63 & 79.81 \\ \hline
300 & Medium & 0.5 & 77.23 & 42.58 & 65.22 & 55.08 & 59.72 & 79.38 & 67.39 & 77.56 & 81.75 & 79.60 \\ \hline
400 & Medium & 0.5 & 78.50 & 44.64 & 66.45 & 57.63 & 61.72 & 79.31 & 67.49 & 77.36 & 82.21 & 79.71 \\ \hline
500 & Medium & 0.5 & 74.85 & 39.47 & 64.73 & 50.28 & 56.60 & 79.30 & 67.44 & 77.13 & 82.47 & 79.71 \\ \hline
\end{tabular}
\vspace{-5mm}
\end{table}

\begin{table}[h!]
\centering
\caption{Performance results for data augmentation of \textit{lb-shortSkirt} in RAPzs}
\label{tab:rapzs_shortskirt}
\setlength{\tabcolsep}{3pt}
\begin{tabular}{c|c|c|c|c|c|c|c|c|c|c|c|c}
\hline
\textbf{\% Augmentation} & \textbf{Noise} & \textbf{Aug. Weight} & \multicolumn{5}{c|}{\textbf{ShortSkirt}} & \multicolumn{5}{c}{\textbf{All Attributes}} \\
\cline{4-13}
 &  &  & \textbf{ma} & \textbf{Acc} & \textbf{Prec} & \textbf{Rec} & \textbf{F1} & \textbf{ma} & \textbf{Acc} & \textbf{Prec} & \textbf{Rec} & \textbf{F1} \\
\hline\hline
50  & Medium & 0.5 & 72.01 & 32.83 & 54.62 & 45.14 & 49.43 & 73.72 & 65.80 & 78.21 & 78.52 & 78.37 \\ \hline
100 & Medium & 0.5 & 70.58 & 30.35 & 51.69 & 42.36 & 46.56 & 73.83 & 65.80 & 77.99 & 78.74 & 78.36 \\ \hline
200 & Medium & 0.5 & 70.14 & 33.33 & 64.13 & 40.97 & 50.00 & 73.77 & 65.84 & 77.82 & 78.95 & 78.38 \\ \hline
300 & Medium & 0.5 & 68.96 & 29.47 & 54.90 & 38.89 & 45.53 & 73.79 & 65.60 & 77.49 & 79.06 & 78.27 \\ \hline
400 & Medium & 0.5 & 69.27 & 29.38 & 53.27 & 39.58 & 45.42 & 73.85 & 65.33 & 76.72 & 79.50 & 78.09 \\ \hline
500 & Medium & 0.5 & 69.11 & 31.82 & 63.64 & 38.89 & 48.28 & 73.81 & 65.60 & 76.78 & 79.83 & 78.28 \\ \hline
\end{tabular}
\end{table}

\begin{table*}[t]
\centering
\caption{Comparison of F1 on the attribute and F1 on the Dataset for the \textit{lb-ShortSkirt} attribute across RAPv1, RAPv2, and RAPzs datasets for different data augmentation (DA) percentages.}
\vspace{-2mm}
\label{tab:all_shortskirt_results}
\setlength{\tabcolsep}{3pt}
\begin{tabular}{c|c|c|c|c|c|c}
\hline
\textbf{DA (\%)} & \multicolumn{2}{|c|}{\textbf{RAPv1}} & \multicolumn{2}{|c|}{\textbf{RAPv2}} & \multicolumn{2}{|c}{\textbf{RAPzs}} \\ \cline{2-7}
                 & \textbf{F1 Attribute} & \textbf{F1 Dataset} & \textbf{F1 Attribute} & \textbf{F1 Dataset} & \textbf{F1 Attribute} & \textbf{F1 Dataset} \\ \hline \hline
50   & 65.57                & 80.50               & 61.65                & 79.39               & 49.43                & 78.37               \\ \hline
100  & 68.51                & 80.52               & 59.35                & 79.48               & 46.56                & 78.36               \\ \hline
200  & 66.11                & 80.51               & 59.54                & \textbf{79.81}      & 50.00                & \textbf{78.38}      \\ \hline
300  & 65.50                & 80.09               & 59.72                & 79.60               & 45.53                & 78.27               \\ \hline
400  & 67.68                & \textbf{80.63}      & 61.72                & 79.71               & 45.42                & 78.09               \\ \hline
500  & 63.49                & 80.53               & 56.60                & 79.71               & 48.28                & 78.28               \\ \hline\hline
Optimized baseline results & \textbf{68.62}   & 80.30               & \textbf{62.54}       & 79.38               & \textbf{51.30}       & 78.20               \\ \hline
\end{tabular}
\end{table*}

\begin{figure}[h!]
    \centering
    \includegraphics[width=0.8\columnwidth]{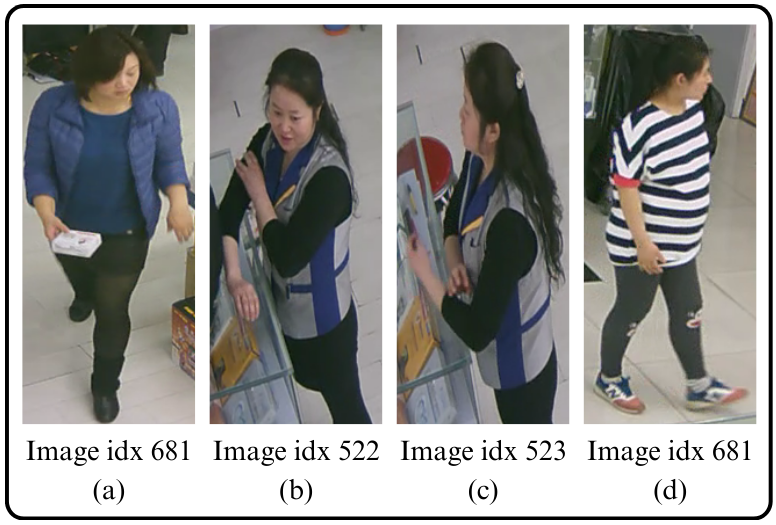}
    \vspace{-2mm}
    \caption{Illustrative examples of challenging cases for the lb-ShortSkirt attribute: (a) and (b) present difficulties due to color similarity, (c) requires contextual inference from (b) as it is ambiguous in isolation, and (d) highlights potential labeling inconsistencies.}
    \vspace{-2mm}
    \label{fig:shortskirt_challenges}
\end{figure}

\newpage
In this case, the problem is unlikely to originate from the testing images per se. Unlike the \textit{hs-BaldHead} scenario in RAPzs, where a limited set of test images exacerbates the impact of noise, the substantially larger test set here contains a smaller percentage of it. While it is present, its influence is discernible yet not critically detrimental. An analysis of the failure cases, as illustrated in Figure \ref{fig:shortskirt_challenges}, corroborates this observation.

A significant portion of the test images for this attribute contain short skirts that are challenging to recognize even for human observers. This increases the difficulty of classification, potentially explaining why the baseline model struggles with this attribute, despite having access to a relatively large number of training samples.

\begin{figure}[b]
    \centering
    \includegraphics[width=0.78\columnwidth]{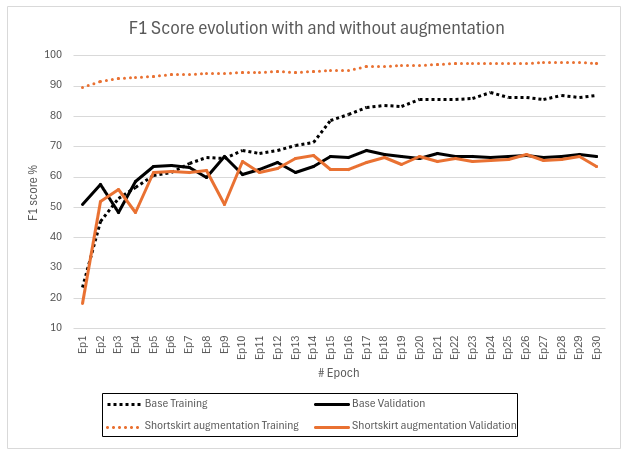}
    \vspace{-5mm}
    \caption{F1 score on lb-ShortSkirt during the different epochs of training for the base model and the augmentated one.}
    \label{plot:shortskirt_learning_plot}
\end{figure}

This hypothesis is further supported by the learning curves (Figure \ref{plot:shortskirt_learning_plot}), which show that, across all augmentations, the model continues to improve its performance on the training set, achieving higher accuracy compared to training without augmentation. This suggests that the model is effectively learning from the synthetic images. However, the observed gap between training and test performance indicates overfitting, implying that the learned representations from augmented data do not generalize well to the test set.

However, it is evident that this augmentation fails to resolve the issue associated with the given attribute. Specifically, generations from a general prompt are insufficient in addressing highly specific problems, leading to the inclusion of broader examples that act as distractors within the dataset, ultimately degrading the attribute-specific scores.

Despite the lack of improvement in the targeted attribute, an overall performance gain is observed across the dataset. Given that the baseline for comparison is the previously computed average, one might expect the results to remain unchanged or even deteriorate. However, contrary to this expectation, a positive trend emerges, for instance, RAPv1 exhibits a two-point improvement.
This suggests that attributes indirectly influenced by label $3$ benefit the model, even though the intended attribute does not generalize well. While the augmentation fails to enhance the specific attribute, it effectively improves the model’s performance on others.

\vspace{-2mm}
\paragraph{Results on AgeLess16:}

The third attribute to test is \textit{AgeLess16}, which differs significantly from the previous two as it is the only one that does not refer to something the person is wearing (such as a hairstyle or a piece of clothing), but rather to something the person is. This distinction makes it particularly challenging, as the boundary between \textit{AgeLess16} and the adjacent class, \textit{Age17-30}, is subtle and difficult to define in practice.

To generate samples for this attribute, the objective was to produce pedestrian images without age ambiguity, avoiding cases that could be misclassified as \textit{Age17-30}. Consequently, the chosen prompt emphasized "kid or teenager clearly not older than 15 years old" to ensure the generated individuals were well outside the ambiguous age range.

Following this approach, all required augmentation samples were synthesized, adhering to the values presented in Table \ref{tab:all_ageLess16_nImages}. In terms of dataset size, this case falls between the previous two: it contains fewer base images than \textit{lb-ShortSkirt}, yet more than \textit{hs-BaldHead}.

\begin{table}[t]
\centering
\captionsetup{justification=centering}
\caption{Number of AgeLess16 train and test images for RAPv1, RAPv2, and RAPzs datasets.}
\label{tab:all_ageLess16_nImages}
\begin{tabular}{c|c|c}
\hline
\textbf{Dataset} & \textbf{\# AgeLess16 Train Images} & \textbf{\# AgeLess16 Test Images} \\ \hline \hline
RAPv1          & 334                                & 81                                 \\ \hline
RAPv2          & 603                                & 150                                \\ \hline
RAPzs          & 195                                & 68                                 \\ \hline
\end{tabular}
\vspace{-8mm}
\end{table}

\begin{figure}[b]
    \centering
    \includegraphics[width=0.82\columnwidth]{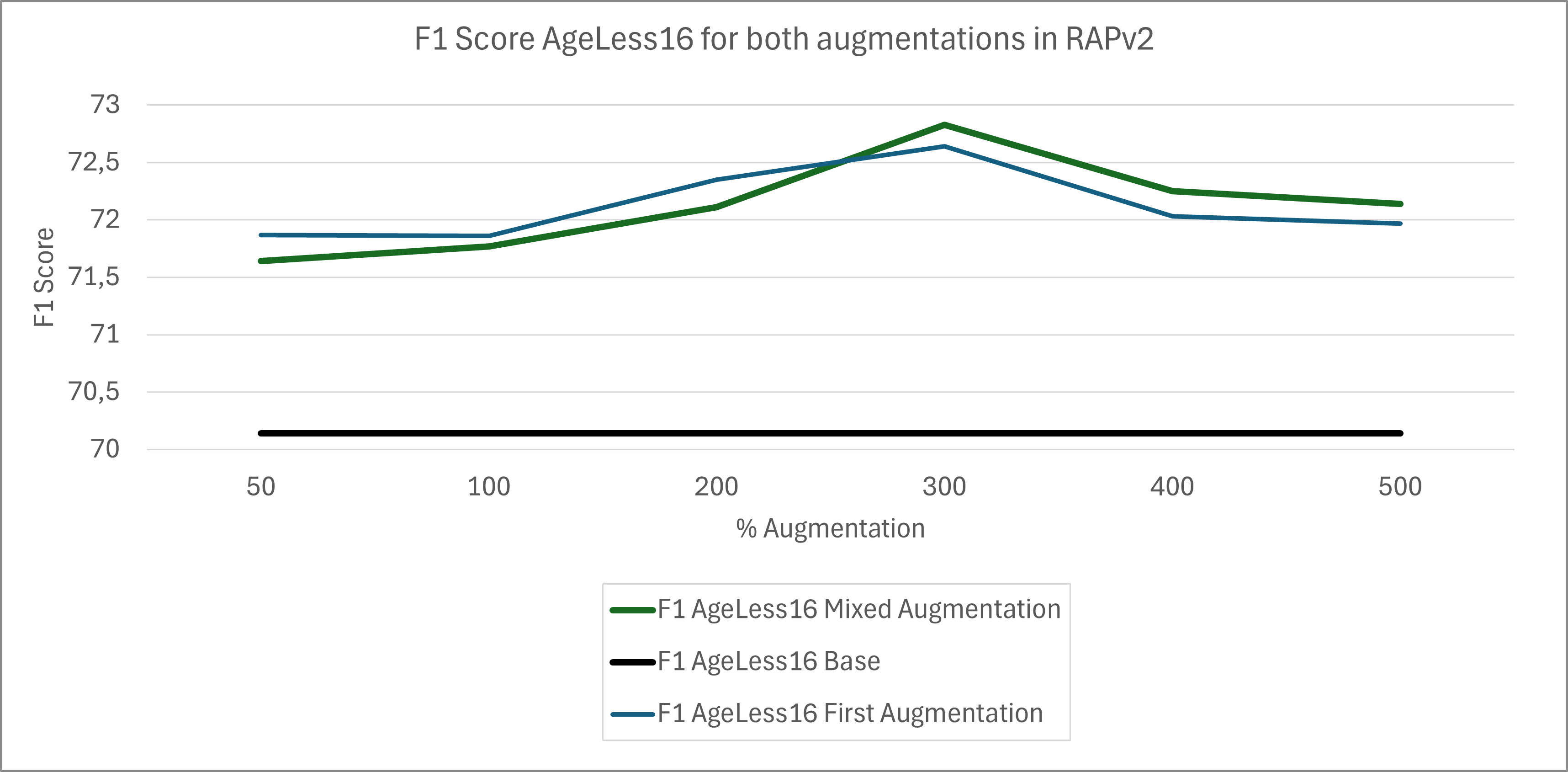}
    \caption{F1 score on RapV2 with the base model, male augmentation (First) and mixed augmentation.}
    \vspace{-5mm}
    \label{plot:f1_ageless16_fyoung_plot}
\end{figure}

After filtering the generated images, a notable observation emerges: despite the prompt containing no explicit reference to pedestrian gender, over 95\% of the generated pedestrians were male. This reveals a significant bias in the generative model with the given prompt.
To mitigate this issue, an additional generation was conducted by explicitly incorporating the term female into the prompt. The newly generated images were then alternated with the original set in a one-to-one manner to enhance generalization.
Both augmentation sets were subsequently utilized for evaluation, as both had been prepared. Surprisingly, when comparing them, the difference is not significant (an example for RAPv2 can be seen in Figure \ref{plot:f1_ageless16_fyoung_plot}). 
This may suggest the absence of female children in the evaluation dataset; however, this is not the case. Instead, the model's failure in these instances appears to be influenced by other underlying factors.

All the experiments performed are presented in Tables \ref{table:rapv1_ageless16}, \ref{table:rapv1_ageless16_female}, \ref{table:rapv2_ageless16}, \ref{table:rapv2_ageless16_female}, \ref{tab:rapzs_ageless16} and \ref{table:rapzs_ageless16_female}. A summary is given in Table \ref{tab:all_ageless16_results}.

\begin{table}[h!]
\centering
\caption{RAPv1 - AgeLess16 Augmentation Results}
\label{table:rapv1_ageless16}
\setlength{\tabcolsep}{3pt}
\begin{tabular}{c|c|c|c|c|c|c|c|c|c|c|c|c}
\hline
\textbf{\% Augmentation} & \textbf{Noise} & \textbf{Aug. Weight} & \multicolumn{5}{c|}{\textbf{AgeLess16}} & \multicolumn{5}{c}{\textbf{All Attributes}} \\
\cline{4-13}
 & & & \textbf{ma} & \textbf{Acc} & \textbf{Prec} & \textbf{Rec} & \textbf{F1} & \textbf{ma} & \textbf{Acc} & \textbf{Prec} & \textbf{Rec} & \textbf{F1} \\
\hline\hline
50  & Medium & 0.5 & 85.12 & 61.96 & 83.82 & 70.37 & 76.51 & 80.36 & 68.49 & 79.95 & 80.75 & 80.35 \\ \hline
100 & Medium & 0.5 & 88.83 & 69.23 & 86.30 & 77.78 & 81.82 & 80.79 & 68.25 & 79.83 & 80.48 & 80.15 \\ \hline
200 & Medium & 0.5 & 85.09 & 58.76 & 78.08 & 70.37 & 74.03 & 80.67 & 68.19 & 79.20 & 80.99 & 80.08 \\ \hline
300 & Medium & 0.5 & 89.38 & 62.75 & 75.29 & 79.01 & 77.11 & 80.80 & 68.50 & 79.39 & 81.27 & 80.32 \\ \hline
400 & Medium & 0.5 & 85.15 & 65.52 & 90.48 & 70.37 & 79.17 & 80.82 & 68.63 & 79.67 & 81.26 & 80.46 \\ \hline
500 & Medium & 0.5 & 85.11 & 61.29 & 82.61 & 70.37 & 76.00 & 81.01 & 68.92 & 79.57 & 81.83 & 80.68 \\ \hline
\end{tabular}
\end{table}

\begin{table}[h!]
\centering
\caption{RAPv1 - AgeLess16 (with Female) Augmentation Results}
\label{table:rapv1_ageless16_female}
\setlength{\tabcolsep}{3pt}
\begin{tabular}{c|c|c|c|c|c|c|c|c|c|c|c|c}
\hline
\textbf{\% Augmentation} & \textbf{Noise} & \textbf{Aug. Weight} & \multicolumn{5}{c|}{\textbf{AgeLess16 (with Female)}} & \multicolumn{5}{c}{\textbf{All Attributes}} \\
\cline{4-13}
 & & & \textbf{ma} & \textbf{Acc} & \textbf{Prec} & \textbf{Rec} & \textbf{F1} & \textbf{ma} & \textbf{Acc} & \textbf{Prec} & \textbf{Rec} & \textbf{F1} \\
\hline\hline
50  & Medium & 0.5 & 83.90 & 61.11 & 85.94 & 67.90 & 75.86 & 80.49 & 68.21 & 79.75 & 80.45 & 80.10 \\ \hline
100 & Medium & 0.5 & 85.73 & 62.37 & 82.86 & 71.60 & 76.82 & 80.63 & 68.34 & 79.88 & 80.57 & 80.22 \\ \hline
200 & Medium & 0.5 & 86.34 & 62.77 & 81.94 & 72.84 & 77.12 & 80.75 & 68.78 & 79.51 & 81.66 & 80.57 \\ \hline
300 & Medium & 0.5 & 89.39 & 64.00 & 77.11 & 79.01 & 78.05 & 80.82 & 68.20 & 79.31 & 80.99 & 80.14 \\ \hline
400 & Medium & 0.5 & 86.36 & 64.84 & 85.51 & 72.84 & 78.67 & 80.98 & 68.70 & 79.60 & 81.45 & 80.51 \\ \hline
500 & Medium & 0.5 & 84.51 & 61.54 & 84.85 & 69.14 & 76.19 & 80.90 & 68.78 & 79.54 & 81.71 & 80.61 \\ \hline
\end{tabular}
\end{table}

\begin{table}[h!]
\centering
\caption{RAPv2 - AgeLess16 Augmentation results. Metrics include mean average (ma), accuracy (Acc), precision (Prec), recall (Rec), and F1 score for AgeLess16 and All Attributes under medium noise conditions with varying augmentation percentages.}
\label{table:rapv2_ageless16}
\setlength{\tabcolsep}{3pt}
\begin{tabular}{c|c|c|c|c|c|c|c|c|c|c|c|c}
\hline
\textbf{\% Augmentation} & \textbf{Noise} & \textbf{Aug. Weight} & \multicolumn{5}{c|}{\textbf{AgeLess16}} & \multicolumn{5}{c}{\textbf{All Attributes}} \\
\cline{4-13}
 &  &  & \textbf{ma} & \textbf{Acc} & \textbf{Prec} & \textbf{Rec} & \textbf{F1} & \textbf{ma} & \textbf{Acc} & \textbf{Prec} & \textbf{Rec} & \textbf{F1} \\
\hline\hline
50  & Medium & 0.5 & 79.77 & 56.09 & 74.21 & 69.68 & 71.87 & 78.86 & 67.39 & 78.59 & 80.53 & 79.54 \\ \hline
100 & Medium & 0.5 & 80.15 & 56.08 & 71.02 & 72.71 & 71.86 & 79.04 & 67.34 & 78.16 & 80.94 & 79.53 \\ \hline
200 & Medium & 0.5 & 80.43 & 56.68 & 71.97 & 72.73 & 72.35 & 79.17 & 67.33 & 78.18 & 80.92 & 79.52 \\ \hline
300 & Medium & 0.5 & 80.91 & 57.03 & 70.38 & 75.04 & 72.64 & 79.42 & 67.48 & 78.05 & 81.32 & 79.65 \\ \hline
400 & Medium & 0.5 & 80.23 & 56.28 & 71.43 & 72.63 & 72.03 & 79.15 & 67.20 & 77.73 & 81.20 & 79.43 \\ \hline
500 & Medium & 0.5 & 80.25 & 56.21 & 70.94 & 73.03 & 71.97 & 79.25 & 67.31 & 77.67 & 81.48 & 79.53 \\ \hline
\end{tabular}
\end{table}

\begin{table}[h!]
\centering
\caption{RAPv2 - AgeLess16 (with Female) Augmentation results}
\label{table:rapv2_ageless16_female}
\setlength{\tabcolsep}{3pt}
\begin{tabular}{c|c|c|c|c|c|c|c|c|c|c|c|c}
\hline
\textbf{\% Augmentation} & \textbf{Noise} & \textbf{Aug. Weight} & \multicolumn{5}{c|}{\textbf{AgeLess16 (with Female)}} & \multicolumn{5}{c}{\textbf{All Attributes}} \\
\cline{4-13}
 &  &  & \textbf{ma} & \textbf{Acc} & \textbf{Prec} & \textbf{Rec} & \textbf{F1} & \textbf{ma} & \textbf{Acc} & \textbf{Prec} & \textbf{Rec} & \textbf{F1} \\
\hline\hline
50  & Medium & 0.5 & 79.65 & 55.81 & 73.60 & 69.78 & 71.64 & 78.62 & 67.27 & 78.53 & 80.44 & 79.47 \\ \hline
100 & Medium & 0.5 & 80.08 & 55.96 & 70.92 & 72.63 & 71.77 & 79.20 & 67.18 & 77.98 & 80.87 & 79.40 \\ \hline
200 & Medium & 0.5 & 80.21 & 56.38 & 72.17 & 72.05 & 72.11 & 79.15 & 67.35 & 78.24 & 80.86 & 79.53 \\ \hline
300 & Medium & 0.5 & 81.09 & 57.27 & 70.34 & 75.50 & 72.83 & 79.40 & 67.62 & 78.16 & 81.40 & 79.75 \\ \hline
400 & Medium & 0.5 & 80.40 & 56.56 & 71.64 & 72.87 & 72.25 & 79.12 & 67.16 & 77.78 & 81.14 & 79.42 \\ \hline
500 & Medium & 0.5 & 80.31 & 56.43 & 71.61 & 72.69 & 72.14 & 79.32 & 67.51 & 77.83 & 81.60 & 79.67 \\ \hline
\end{tabular}
\end{table}

\begin{table}[h!]
\centering
\caption{RAPzs - AgeLess16 Augmentation results. Metrics include mean average (ma), accuracy (Acc), precision (Prec), recall (Rec), and F1 score for AgeLess16 and All Attributes under medium noise conditions with varying augmentation percentages.}
\label{tab:rapzs_ageless16}
\setlength{\tabcolsep}{3pt}
\begin{tabular}{c|c|c|c|c|c|c|c|c|c|c|c|c}
\hline
\textbf{\% Augmentation} & \textbf{Noise} & \textbf{Aug. Weight} & \multicolumn{5}{c|}{\textbf{AgeLess16}} & \multicolumn{5}{c}{\textbf{All Attributes}} \\
\cline{4-13}
 & & & \textbf{ma} & \textbf{Acc} & \textbf{Prec} & \textbf{Rec} & \textbf{F1} & \textbf{ma} & \textbf{Acc} & \textbf{Prec} & \textbf{Rec} & \textbf{F1} \\
\hline\hline
50  & Medium & 0.5 & 64.62 & 26.32 & 71.43 & 29.41 & 41.67 & 72.95 & 65.34 & 78.42 & 77.51 & 77.96 \\ \hline
100 & Medium & 0.5 & 70.48 & 35.44 & 71.79 & 41.18 & 52.34 & 73.16 & 65.78 & 78.56 & 78.08 & 78.32 \\ \hline
200 & Medium & 0.5 & 70.53 & 37.84 & 82.35 & 41.18 & 54.90 & 73.50 & 65.24 & 77.66 & 78.17 & 77.91 \\ \hline
300 & Medium & 0.5 & 69.76 & 35.06 & 75.00 & 39.71 & 51.92 & 73.53 & 65.19 & 77.74 & 78.15 & 77.94 \\ \hline
400 & Medium & 0.5 & 70.43 & 33.73 & 65.12 & 41.18 & 50.45 & 74.03 & 65.81 & 77.85 & 78.97 & 78.40 \\ \hline
500 & Medium & 0.5 & 74.14 & 41.25 & 73.33 & 48.53 & 58.41 & 74.19 & 65.75 & 77.74 & 78.95 & 78.34 \\ \hline
\end{tabular}
\end{table}

\begin{table}[t]
\centering
\caption{RAPzs - AgeLess16 (with Female) Augmentation Results}
\label{table:rapzs_ageless16_female}
\setlength{\tabcolsep}{3pt}
\begin{tabular}{c|c|c|c|c|c|c|c|c|c|c|c|c}
\hline
\textbf{\% Augmentation} & \textbf{Noise} & \textbf{Aug. Weight} & \multicolumn{5}{c|}{\textbf{AgeLess16 (with Female)}} & \multicolumn{5}{c}{\textbf{All Attributes}} \\
\cline{4-13}
 & & & \textbf{ma} & \textbf{Acc} & \textbf{Prec} & \textbf{Rec} & \textbf{F1} & \textbf{ma} & \textbf{Acc} & \textbf{Prec} & \textbf{Rec} & \textbf{F1} \\
\hline\hline
50  & Medium & 0.5 & 65.39 & 28.77 & 80.77 & 30.88 & 44.68 & 73.06 & 65.37 & 78.50 & 77.51 & 78.00 \\ \hline
100 & Medium & 0.5 & 67.56 & 31.58 & 75.00 & 35.29 & 48.00 & 72.95 & 65.60 & 78.36 & 77.96 & 78.16 \\ \hline
200 & Medium & 0.5 & 71.97 & 38.96 & 76.92 & 44.12 & 56.07 & 73.67 & 65.09 & 77.48 & 78.13 & 77.80 \\ \hline
300 & Medium & 0.5 & 73.44 & 41.56 & 78.05 & 47.06 & 58.72 & 74.12 & 65.64 & 77.73 & 78.85 & 78.29 \\ \hline
400 & Medium & 0.5 & 69.78 & 36.00 & 79.41 & 39.71 & 52.94 & 73.98 & 65.95 & 77.94 & 79.08 & 78.51 \\ \hline
500 & Medium & 0.5 & 74.10 & 39.29 & 67.35 & 48.53 & 56.41 & 74.14 & 65.74 & 77.69 & 78.99 & 78.33 \\ \hline
\end{tabular}
\end{table}

In any case, the general results are presented using the augmentation set that includes \textit{female kids}, as it tends to yield similar results but is considered more general. These results can be found in Table \ref{tab:all_ageless16_results}.
In this case, it is worth highlighting that although the baseline results are relatively high compared to the previous attributes, and thus, one might expect them to remain similar or show only minor improvements, several points of improvement are achieved in the metrics for both RAPv1 and RAPv2. 
Furthermore, regarding the score for the entire dataset, it is notable that the trend of consistently improving it continues. Therefore, overall, the approach also proves to be effective in this case.
However, it is important to highlight the situation for the attribute in RAPzs. Unlike the previous two datasets, the attribute itself does not improve, and in some instances, it even worsens considerably. This is notable, particularly given that the same augmentations did yield better results in RAPv1 and RAPv2. 
This indicates that part of those improved scores are there because the data leakage inherent in the construction of these training and testing sets is behind, as noted by Jia et al. \citesupp{jia2021rethinking}. Additionally, it implies that the augmentation is unable to reproduce the patterns now missing due to the loss of duplicated identities across both subsets, which points to certain characteristics that are difficult to replicate. 

\begin{table*}[t]
\centering
\caption{Comparison of F1 on the Attribute and F1 on the Dataset for the \textit{AgeLess16} attribute across RAPv1, RAPv2, and RAPzs datasets for different data augmentation (DA) percentages.}
\vspace{-2mm}
\label{tab:all_ageless16_results}
\setlength{\tabcolsep}{3pt}
\begin{tabular}{c|c|c|c|c|c|c}
\hline
\textbf{DA} (\%) & \multicolumn{2}{|c|}{\textbf{RAPv1}} & \multicolumn{2}{|c|}{\textbf{RAPv2}} & \multicolumn{2}{|c}{\textbf{RAPzs}} \\ \cline{2-7}
                 & \textbf{F1 Attribute} & \textbf{F1 Dataset} & \textbf{F1 Attribute} & \textbf{F1 Dataset} & \textbf{F1 Attribute} & \textbf{F1 Dataset} \\ \hline \hline
50   & 75.86                & 80.10               & 71.64                & 79.47               & 44.68                & 78.00               \\ \hline
100  & 76.82                & 80.22               & 71.77                & 79.40               & 48.00                & 78.16               \\ \hline
200  & 77.12                & 80.57               & 72.11                & 79.53               & 56.07                & 77.80               \\ \hline
300  & 78.05                & 80.14               & \textbf{72.83}       & \textbf{79.75}      & 58.72                & 78.29               \\ \hline
400  & \textbf{78.67}       & 80.51               & 72.25                & 79.42               & 52.94                & \textbf{78.51}      \\ \hline
500  & 76.19                & \textbf{80.61}      & 72.14                & 79.67               & 56.41                & 78.33               \\ \hline \hline
Optimized baseline results & 75.82          & 80.30               & 70.14                & 79.38               & \textbf{59.02}       & 78.20               \\ \hline
\end{tabular}
\vspace{-2mm}
\end{table*}

\begin{figure}[b]
    \centering
    \includegraphics[width=0.60\columnwidth]{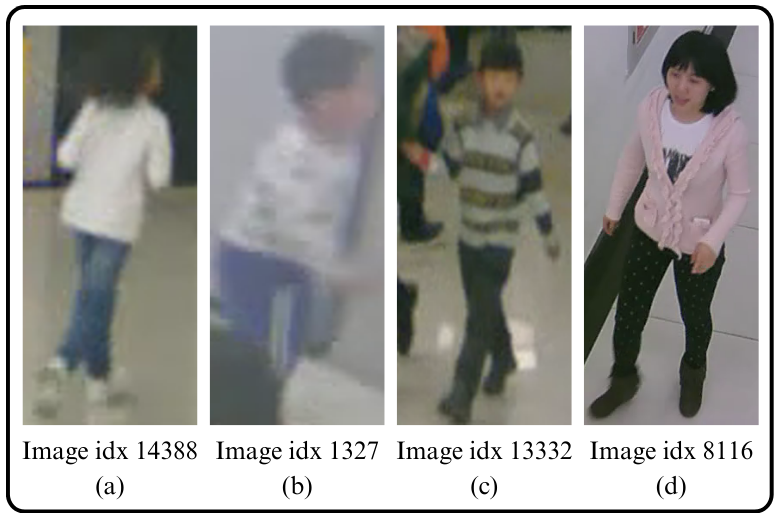}
    \vspace{-2mm}
    \caption{Examples of challenging cases for the AgeLess16 attribute: (a), (b), and (c) depict individuals with poor image quality due to resizing, showing back, side, and front views respectively, where (b) and (c) lose facial details. (d) shows a pedestrian with ambiguous age characteristics.}
    \vspace{-2mm}
    \label{fig:ageless16_challenges}
\end{figure}

By further exploration of the failing cases in the dataset, it is observed that a significant portion of the cases where the classification fails is due to a phenomenon that had not arisen until now. Since children tend to be smaller, the bounding boxes generated around them from the same camera angles are smaller compared to those for adults. Additionally, in many cases, children are not close to the camera, resulting in very small base images that, while interpretable, lose much of the fine-grained details due to image quality, and degrading one of the main features with which we can differentiate the age of a person; the face (Figure \ref{fig:ageless16_challenges}). 
As a result, in these cases, the model must rely solely on body shape to determine age, which appears to present some challenges. Furthermore, the model also fails in cases where children are viewed from behind, suggesting a similar issue. This observation makes sense considering the image generation process. In general, generative models often struggle with anatomy and proportions (evident in frequent artifacts like extra or overly long limbs). Although the worst examples of these issues are filtered out during the manual analysis of the generated images, some cases remain where the images, while clearly depicting children based on facial features, have body shapes that might resemble those of adults or fall somewhere in between. 

This finding makes classification based on body shape a challenge, especially given the wide range of body types within the age category, encompassing infants, young children, preteens, and teenagers.
Regulating this issue with text-to-image generation is challenging, so further improvements for this attribute using this approach depend heavily on the generative precision of the model. Nonetheless, given that results are improved in the first two datasets, it can be concluded that no noise is being introduced into the dataset in any case.

\vspace{-2mm}
\paragraph{Results on ub-SuitUp:}

Following the unsuccessful outcomes with lb-ShortSkirt, ub-SuitUp becomes an attribute of particular interest, as it features an even higher number of training images, as can be seen in Table \ref{tab:all_suitup_nImages}.

\begin{table}[t]
\centering
\captionsetup{justification=centering}
\caption{Number of ub-SuitUp train and test images for RAP datasets.}
\label{tab:all_suitup_nImages}
\vspace{-2mm}
\begin{tabular}{c|c|c}
\hline
\textbf{Dataset} & \textbf{\# hs-SuitUp Train Images} & \textbf{\# hs-SuitUp Test Images} \\ 
\hline \hline
RAPv1 & 716 & 205 \\ \hline
RAPv2 & 2067 & 525 \\ \hline
RAPzs & 461 & 112 \\ \hline
\end{tabular}
\end{table}

\begin{table*}[t]
\centering
\caption{Comparison of F1 on the Attribute and F1 on the Dataset for the \textit{ub-SuitUp} attribute across RAPv1, RAPv2, and RAPzs datasets for different data augmentation (DA) percentages.}
\label{tab:all_suitup_results}
\setlength{\tabcolsep}{5pt}
\begin{tabular}{c|c|c|c|c|c|c}
\hline
\textbf{DA} & \multicolumn{2}{c|}{\textbf{RAPv1}} & \multicolumn{2}{c|}{\textbf{RAPv2}} & \multicolumn{2}{c}{\textbf{RAPzs}} \\ 
\cline{2-7}
(\%) & \textbf{F1 Attribute} & \textbf{F1 Dataset} 
 & \textbf{F1 Attribute} & \textbf{F1 Dataset}
 & \textbf{F1 Attribute} & \textbf{F1 Dataset} \\ 
\hline\hline
50   & 67.70 & 80.14 & 65.50 & 79.72 & 64.25 & \textbf{78.46} \\ \hline
100  & 68.42 & 80.41 & \textbf{66.87} & 79.55 & 67.94 & 78.19 \\ \hline
200  & 67.02 & 80.42 & 64.55 & 79.64 & 63.68 & 78.09 \\ \hline
300  & 66.99 & 80.29 & 65.07 & 79.63 & 66.02 & 78.31 \\ \hline
400  & 68.25 & \textbf{80.66} & 66.00 & 79.62 & 65.37 & 78.19 \\ \hline
500  & 66.30 & 80.45 & 63.19 & \textbf{79.79} & \textbf{68.32} & 78.32 \\ \hline
\hline
Optimized baseline results & \textbf{69.29} & 80.30 & 65.83 & 79.38 & 66.67 & 78.20 \\
\hline
\end{tabular}
\end{table*}

Following the experiments, our findings reveal that the specific attribute in question has indeed been enhanced (Table \ref{tab:all_suitup_results}). This outcome contrasts with the behavior observed in the lb-ShortSkirt case, where such improvement was absent, thereby eliminating the possibility that the number of images directly influenced the obtained results. However, in the case of ub-SuitUp, a novel behavior emerges: while both RAPv2 and RAPzs improve, RAPv1 experiences a decline. Given that RAPv2 encompasses RAPv1, the decrease in RAPv1 is of lesser concern. Although pinpointing why improvements are isolated to RAPv2 and RAPzs remains challenging, even if it proves that the proposed approach does not introduce noise. Moreover, the enhancement observed in the zero-shot variant of the datasets is particularly significant; it deviates from the patterns seen with previous attributes and indicates that incorporating synthetic data enhances generalization under these conditions.
When evaluating the metrics on the complete dataset, we can observe again consistent improvements. These gains are more pronounced at higher augmentation percentages, although peaks in performance may occur at specific augmentation levels.

\paragraph{Results on attach-PaperBag:}
The attribute attach-PaperBag exhibits a number of images comparable to most attributes in the dataset (see Table \ref{tab:all_paperbag_nImages}); however, its baseline performance remains considerably low. This behavior may be attributed to its similarity to other attributes in the same category, particularly attach-PlasticBag.

\begin{table}[t]
\centering
\captionsetup{justification=centering}
\caption{Number of attach-PaperBag train and test images for RAP datasets.}
\label{tab:all_paperbag_nImages}
\vspace{-2mm}
\begin{tabular}{c|c|c}
\hline
\textbf{Dataset} & \textbf{\# attach-PaperBag Train Images} & \textbf{\# attach-PaperBag Test Images} \\ 
\hline \hline
RAPv1 & 397 & 129 \\ \hline
RAPv2 & 704 & 181 \\ \hline
RAPzs & 197 & 54 \\ \hline
\end{tabular}
\end{table}

As can be seen, our experiments on the attribute (Table \ref{tab:all_paperbag_results}) demonstrates a compelling use-case of the proposed approach. By integrating data augmentation within the attribute module, we observe notable performance gains on both the official datasets and their zero-shot variants. These improvements are evident at both the attribute and dataset levels, underscoring the efficacy of the method.

Moreover, the attribute recognition results on RAPzs capture one of the primary goals of our work: enhancing the detection of underrepresented attributes in zero-shot scenarios.

\begin{table*}[t]
\centering
\caption{Comparison of F1 on the Attribute and F1 on the Dataset for the \textit{attach-PaperBag} attribute across RAPv1, RAPv2, and RAPzs datasets for different data augmentation (DA) percentages.}
\label{tab:all_paperbag_results}
\setlength{\tabcolsep}{3pt}
\begin{tabular}{c|c|c|c|c|c|c}
\hline
\textbf{DA} & \multicolumn{2}{c|}{\textbf{RAPv1}} & \multicolumn{2}{c|}{\textbf{RAPv2}} & \multicolumn{2}{c}{\textbf{RAPzs}} \\
\cline{2-7}
(\%) & \textbf{F1 Attribute} & \textbf{F1 Dataset}
 & \textbf{F1 Attribute} & \textbf{F1 Dataset}
 & \textbf{F1 Attribute} & \textbf{F1 Dataset} \\
\hline\hline
50   & 51.35 & 80.17 & 41.48 & 79.52 & 17.72 & 78.21 \\ \hline
100  & 51.10 & 80.45 & 41.09 & 79.62 & 20.93 & 78.27 \\ \hline
200  & 49.54 & 80.33 & 40.00 & 79.79 & 25.88 & 78.10 \\ \hline
300  & \textbf{54.17} & 80.27 & 41.64 & 79.75 & \textbf{33.33} & 78.04 \\ \hline
400  & 45.37 & \textbf{80.52} & 44.14 & \textbf{79.80} & 30.93 & \textbf{78.35} \\ \hline
500  & 49.28 & 80.43 & \textbf{45.14} & 79.63 & 23.38 & 78.19 \\ \hline
\hline
Optimized baseline results & 45.79 & 80.30 & 42.90 & 79.38 & 9.30  & 78.20 \\
\hline
\end{tabular}
\vspace{-5mm}
\end{table*}

\vspace{-5mm}
\paragraph{Overall Performance:}
After optimizing augmentation per attribute, we tested simultaneous augmentation across datasets. We evaluated two models: one using 500\% augmentation for all attributes and another applying each attribute's optimal augmentation level. Since \textit{lb-ShortSkirt} showed no improvement, we also tested augmentation with only \textit{hs-BaldHead} and \textit{AgeLess16}. Table \ref{tab:all_together_results} presents the average results over 10 runs.

\begin{table*}[h!]
\centering
\caption{Performance comparison with updated results over RAPv1. RAPv2. and RAPzs.}
\resizebox{\textwidth}{!}{%
\begin{tabular}{@{}l|c|c|c|c|c|c|c|c|c|c|c|c|c|c|c|cc|c|c|c|c@{}}
\hline
\multirow{3}{*}{\textbf{Condition}} & \multirow{2}{*}{\textbf{\% Aug.}} & \multicolumn{5}{|c|}{\textbf{RAPv1}} & \multicolumn{5}{|c|}{\textbf{RAPv2}} & \multicolumn{5}{|c}{\textbf{RAPzs}} \\ 
\cline{3-7} \cline{8-12} \cline{13-17}
& & \textbf{mA} & \textbf{Acc} & \textbf{Prec} & \textbf{Rec} & \textbf{F1} 
  & \textbf{mA} & \textbf{Acc} & \textbf{Prec} & \textbf{Rec} & \textbf{F1}
  & \textbf{mA} & \textbf{Acc} & \textbf{Prec} & \textbf{Rec} & \textbf{F1} \\ 
\hline
\hline

\multirow{4}{*}{\parbox{2cm}{\raggedright \textbf{Augmenting hs-BaldHead \& AgeLess16}}}

& \multirow{2}{*}{500\% Augmentation}
  & 80.94 & \textbf{68.66} & 79.14 & 81.93 & 80.51
  & 79.45 & 67.33 & 77.31 & 82.01 & 79.59
  & 73.65 & 65.50 & 77.38 & 78.95 & 78.16 \\
& 
  & \scriptsize±0.18 & \textbf{\scriptsize±0.14} & \scriptsize±0.06 & \scriptsize±0.23 & \scriptsize±0.11
  & \scriptsize±0.07 & \scriptsize±0.10 & \scriptsize±0.13 & \scriptsize±0.11 & \scriptsize±0.07
  & \scriptsize±0.18 & \scriptsize±0.13 & \scriptsize±0.14 & \scriptsize±0.13 & \scriptsize±0.11 \\

\cline{2-17}

& \multirow{2}{*}{Best F1 Augmentation}
  & 80.79 & 68.52 & 79.15 & 81.64 & 80.38
  & 79.13 & \textbf{67.63} & 77.86 & 81.78 & \textbf{79.77}
  & 73.65 & 65.50 & 77.38 & 78.95 & 78.16 \\
& 
  & \scriptsize±0.18 & \scriptsize±0.12 & \scriptsize±0.12 & \scriptsize±0.11 & \scriptsize±0.08
  & \scriptsize±0.07 & \textbf{\scriptsize±0.09} & \scriptsize±0.13 & \scriptsize±0.12 & \textbf{\scriptsize±0.07}
  & \scriptsize±0.18 & \scriptsize±0.13 & \scriptsize±0.14 & \scriptsize±0.13 & \scriptsize±0.11 \\

\hline

\multirow{4}{*}{\textbf{Augmenting all}}

& \multirow{2}{*}{500\% Augmentation}
  & \textbf{81.14} & 68.64 & 77.90 & \textbf{83.51} & \textbf{80.60}
  & \textbf{79.75} & 67.27 & 75.87 & \textbf{83.87} & 79.67
  & \textbf{74.57} & \textbf{65.93} & 75.89 & \textbf{81.56} & \textbf{78.62} \\
&
  & \textbf{\scriptsize±0.12} & \scriptsize±0.11 & \scriptsize±0.09 & \textbf{\scriptsize±0.21} & \textbf{\scriptsize±0.09}
  & \textbf{\scriptsize±0.12} & \scriptsize±0.11 & \scriptsize±0.12 & \textbf{\scriptsize±0.08} & \scriptsize±0.08
  & \textbf{\scriptsize±0.16} & \textbf{\scriptsize±0.14} & \scriptsize±0.14 & \textbf{\scriptsize±0.30} & \textbf{\scriptsize±0.11} \\
  
\cline{2-17}

& \multirow{2}{*}{Best F1 Augmentation}
  & 81.01 & 68.63 & 78.80 & 82.29 & 80.50
  & 79.67 & 67.47 & 76.51 & 83.30 & 79.76
  & 73.89 & \textbf{65.93} & 76.95 & 80.16 & 78.52 \\
&
  & \scriptsize±0.18 & \scriptsize±0.16 & \scriptsize±0.11 & \scriptsize±0.14 & \scriptsize±0.11
  & \scriptsize±0.12 & \scriptsize±0.10 & \scriptsize±0.07 & \scriptsize±0.15 & \scriptsize±0.08
  & \scriptsize±0.13 & \textbf{\scriptsize±0.12} & \scriptsize±0.09 & \scriptsize±0.21 & \scriptsize±0.10 \\

\hline
\hline

\multirow{2}{*}{\textbf{Optimized baseline results}}
& -
  & 80.54 & 68.49 & \textbf{79.89} & 80.72 & 80.30
  & 79.12 & 67.17 & \textbf{78.35} & 80.43 & 79.38
  & 74.39 & 65.59 & \textbf{78.19} & 78.20 & 78.20 \\
&
  & \scriptsize±0.17 & \scriptsize±0.10 & \textbf{\scriptsize±0.08} & \scriptsize±0.13 & \scriptsize±0.08
  & \scriptsize±0.10 & \scriptsize±0.10 & \textbf{\scriptsize±0.09} & \scriptsize±0.12 & \textbf{\scriptsize±0.09}
  & \scriptsize±0.22 & \scriptsize±0.23 & \textbf{\scriptsize±0.15} & \scriptsize±0.27 & \scriptsize±0.19 \\

\hline
\end{tabular}%
}
\label{tab:all_together_results}
\end{table*}

As observed, even when directly augmenting only 2 or 3 attributes out of 52, and indirectly enhancing a small additional subset through the proposed augmented loss, results improved across all datasets. 

It is noteworthy that including lb-ShortSkirt in the augmentations tends to be beneficial, as this variation either achieves the best results or very competitive ones. Conversely, omitting it occasionally seems to result in losing the extra boost provided by the indirectly augmented attributes from those images.

Additionally, precision is the only metric that does not show improvement, suggesting that the model exhibits a tendency to classify the augmented classes as positive more frequently, even in cases of incorrect predictions. This aligns with the challenges observed in fully resembling the dataset. Nevertheless, the overall improvement is indisputable, indicating that a generation approach tailored to address specific challenges within the dataset, and/or the augmentation of most or all attributes rather than a limited subset, can lead to significant improvements.

\newpage
\section{ComfyUI configuration} 
\label{annex:ComfyUIDetails}

\subsection{ComfyUI - Generation Process Description}
For the generation process, once the flow was constructed, the decision was made to ensure that all generations were replicable. This involved using the same seed, "123456789," across the \textit{KSampler}, \textit{NoiseGenerator}, and \textit{PromptSampler}, incrementing it for each subsequent generation. However, it appears that the \textit{PromptSampler}, being a custom node, does not handle this functionality correctly, and the seed remains static unless set to random mode. \\


For this reason, the procedure has been maintained for the \textit{KSampler} and \textit{NoiseGenerator}, while the \textit{PromptSampler} has been left in random mode. Nonetheless, all the generated prompts have been recorded, ensuring that, with these prompts and the corresponding seeds, the images can be reproduced. With this setup, the generations were carried out. For each attribute, more images than necessary were always generated, as those that either did not contain the attribute or exhibited significant generation errors were manually discarded. 
For reproducibility, the index of all discarded images have also been saved, which represent a greater or lesser percentage depending on the attribute. \\

Finally, it is important to note that the decision was made to perform the generation process in a distributed manner. Throughout this project, more than double the number of images specified in this report had to be generated. \\

To maintain the reproducibility specified earlier, the seeds were recalculated for each batch. The first image of the first batch always started with the seed \texttt{123456789}, and the seed for the first image of each subsequent batch was calculated by adding the batch size to the batch number, according to Equation \ref{eq:seed_calculation}. \\

\begin{equation}
\label{eq:seed_calculation}
\text{Seed}_{\text{batch}, 1} = \text{Seed}_{\text{initial}} + (\text{Batch Size} \times (\text{Batch Number} - 1))
\end{equation}

\subsection{ComFyUI Workflow Details}

The following describes the \textit{ComfyUI} flow used for image generation. The various nodes are grouped into modules, with their inputs (if applicable) and parameters specified.\newline

\textbf{LOAD MODELS}

\begin{itemize}[label=--]
    \setlength{\itemsep}{0pt}
    \item \textbf{Load Checkpoint}
    \begin{itemize}[label={}]
        \setlength{\itemsep}{0pt}
        \item \texttt{parameters:}
        \begin{itemize}[label={}]
            \setlength{\itemsep}{0pt}
            \item \texttt{ckpt\_name = sd3\_medium\_incl\_clips \_t5xxlfp8.safetensors}
        \end{itemize}
    \end{itemize}
    \item \textbf{TripleCLIPLoader}
    \begin{itemize}[label={}]
        \setlength{\itemsep}{0pt}
        \item \texttt{parameters:}
        \begin{itemize}[label={}]
            \setlength{\itemsep}{0pt}
            \item \texttt{clip\_name1 = clip\_g.safetensors}
            \item \texttt{clip\_name2 = clip\_l.safetensors}
            \item \texttt{clip\_name3 = t5xxl\_lp16.safetensors}
        \end{itemize}
    \end{itemize}
\end{itemize}

\textbf{INPUT}

\begin{itemize}[label=--]
    \setlength{\itemsep}{0pt}
    \item \textbf{EmptySD3LatentImage}
    \begin{itemize}[label={}]
        \setlength{\itemsep}{0pt}
        \item \texttt{parameters:}
        \begin{itemize}[label={}]
            \setlength{\itemsep}{0pt}
            \item \texttt{width = 2784}
            \item \texttt{height = 1024}
            \item \texttt{batch\_size = 1}
        \end{itemize}
    \end{itemize}
    \item \textbf{Random Prompts}  \citesupp{DynamicPrompts}
    \begin{itemize}[label={}]
        \item \texttt{parameters:}
        \begin{itemize}[label={}]
            \item \texttt{text = <prompt>}
            \item \texttt{seed = 123456789}
            \item \texttt{control\_after\_generate = randomize}
            \item \texttt{autorefresh = No}
        \end{itemize}
    \end{itemize}
    \item \textbf{CLIP Text Encode (Prompt)}
    \begin{itemize}[label={}]
        \item \texttt{inputs:}
        \begin{itemize}[label={}]
            \item \texttt{clip = TripleCLIPLoader}
            \item \texttt{text = Random Prompts}
        \end{itemize}
    \end{itemize}
    \item \textbf{CLIP Text Encode (Negative Prompt)}
    \begin{itemize}[label={}]
        \item \texttt{inputs:}
        \begin{itemize}[label={}]
            \item \texttt{clip = TripleCLIPLoader}
        \end{itemize}
        \item \texttt{parameters:}
        \begin{itemize}[label={}]
            \item \texttt{text = <negative prompt>}
        \end{itemize}
    \end{itemize}
    \item \textbf{Show Text Value}
    \begin{itemize}[label={}]
        \item \texttt{inputs:}
        \begin{itemize}[label={}]
            \item \texttt{text = Random Prompts}
        \end{itemize}
        \item \texttt{parameters:}
        \begin{itemize}[label={}]
            \item \texttt{split\_lines = false}
        \end{itemize}
    \end{itemize}
    \item \textbf{Save Text}
    \begin{itemize}[label={}]
        \item \texttt{inputs:}
        \begin{itemize}[label={}]
            \item \texttt{text = Show Text Value}
        \end{itemize}
        \item \texttt{parameters:}
        \begin{itemize}[label={}]
            \item \texttt{filename\_prefix = 1\_ComfyUI}
            \item \texttt{images\_paths = None}
        \end{itemize}
    \end{itemize}
\end{itemize}

\textbf{GENERATION STEPS}

\begin{itemize}[label=--]
    \item \textbf{ModelSamplingSD3}
    \begin{itemize}[label={}]
        \item \texttt{inputs:}
        \begin{itemize}[label={}]
            \item \texttt{model = Load Checkpoints}
        \end{itemize}
        \item \texttt{parameters:}
        \begin{itemize}[label={}]
            \item \texttt{shift = 3.00}
        \end{itemize}
    \end{itemize}
    \item \textbf{ConditioningZeroOut}
    \begin{itemize}[label={}]
        \item \texttt{inputs:}
        \begin{itemize}[label={}]
            \item \texttt{conditioning = CLIP Text Encode (Negative Prompt)}
        \end{itemize}
    \end{itemize}
    \item \textbf{ConditioningSetTimestepRange}
    \begin{itemize}[label={}]
        \item \texttt{inputs:}
        \begin{itemize}[label={}]
            \item \texttt{conditioning = CLIP Text Encode (Negative Prompt)}
        \end{itemize}
        \item \texttt{parameters:}
        \begin{itemize}[label={}]
            \item \texttt{start = 0.000}
            \item \texttt{end = 0.100}
        \end{itemize}
    \end{itemize}
    \item \textbf{ConditioningSetTimestepRange}
    \begin{itemize}[label={}]
        \item \texttt{inputs:}
        \begin{itemize}[label={}]
            \item \texttt{conditioning = ConditioningZeroOut}
        \end{itemize}
        \item \texttt{parameters:}
        \begin{itemize}[label={}]
            \item \texttt{start = 0.100}
            \item \texttt{end = 1.000}
        \end{itemize}
    \end{itemize}
    \item \textbf{Conditioning (Combine)}
    \begin{itemize}[label={}]
        \item \texttt{inputs:}
        \begin{itemize}[label={}]
            \item \texttt{conditioning\_1 = ConditioningSetTimestepRange}
            \item \texttt{conditioning\_2 = ConditioningSetTimestepRange}
        \end{itemize}
    \end{itemize}
    \item \textbf{KSampler}
    \begin{itemize}[label={}]
        \item \texttt{inputs:}
        \begin{itemize}[label={}]
            \item \texttt{model = ModelSamplingSD3}
            \item \texttt{positive = CLIP Text Encode (Prompt)}
            \item \texttt{negative = Conditioning (Combine)}
            \item \texttt{latent\_image = EmptySD3LatentImage}
        \end{itemize}
        \item \texttt{parameters:}
        \begin{itemize}[label={}]
            \item \texttt{seed = 123456789}
            \item \texttt{control\_after\_generate = increment}
            \item \texttt{steps = 28}
            \item \texttt{cfg = 4.5}
            \item \texttt{sampler\_name = dpmpp\_2m}
            \item \texttt{scheduler = sgm\_uniform}
            \item \texttt{denoise = 1.00}
        \end{itemize}
    \end{itemize}
    \item \textbf{VAE Decode}
    \begin{itemize}[label={}]
        \item \texttt{inputs:}
        \begin{itemize}[label={}]
            \item \texttt{samples = KSampler}
            \item \texttt{vae = Load Checkpoint}
        \end{itemize}
    \end{itemize}
\end{itemize}

\textbf{POSTPROCESSING}

\begin{itemize}[label=--]
    \item \textbf{ToolYoloCropper} \citesupp{ComfyUI-Yolo-Cropper}, \citesupp{YoloV8}
    \begin{itemize}[label={}]
        \item \texttt{inputs:}
        \begin{itemize}[label={}]
            \item \texttt{image = VAE Decode}
        \end{itemize}
        \item \texttt{parameters:}
        \begin{itemize}[label={}]
            \item \texttt{object = person}
            \item \texttt{padding = 0}
        \end{itemize}
    \end{itemize}
    \item \textbf{Greyscale Noise} \citesupp{comfy-plasma}
    \begin{itemize}[label={}]
        \item \texttt{parameters:}
        \begin{itemize}[label={}]
            \item \texttt{width = 1024}
            \item \texttt{height = 1024}
            \item \texttt{value\_min = -1}
            \item \texttt{value\_max = -1}
            \item \texttt{red\_min = -1}
            \item \texttt{red\_max = -1}
            \item \texttt{green\_min = -1}
            \item \texttt{green\_max = -1}
            \item \texttt{blue\_min = -1}
            \item \texttt{blue\_max = -1}
            \item \texttt{seed = 123456789}
            \item \texttt{control\_after\_generate = increment}
        \end{itemize}
    \end{itemize}
    \item \textbf{ImageBlend}
    \begin{itemize}[label={}]
        \item \texttt{inputs:}
        \begin{itemize}[label={}]
            \item \texttt{image1 = ToolYoloCropper}
            \item \texttt{image2 = Greyscale Noise}
        \end{itemize}
        \item \texttt{parameters:}
        \begin{itemize}[label={}]
            \item \texttt{blend\_factor = 0.25}
            \item \texttt{blend\_mode = soft\_light}
        \end{itemize}
    \end{itemize}
    \item \textbf{Upscale Image By} \citesupp{image-resize-comfyui} 
    \begin{itemize}[label={}]
        \item \texttt{inputs:}
        \begin{itemize}[label={}]
            \item \texttt{image = ImageBlend}
        \end{itemize}
        \item \texttt{parameters:}
        \begin{itemize}[label={}]
            \item \texttt{upscale\_method = nearest-exact}
            \item \texttt{scale\_by = 0.33}
        \end{itemize}
    \end{itemize}
    \item \textbf{Upscale Image By}
    \begin{itemize}[label={}]
        \item \texttt{inputs:}
        \begin{itemize}[label={}]
            \item \texttt{image = Upscale Image By}
        \end{itemize}
        \item \texttt{parameters:}
        \begin{itemize}[label={}]
            \item \texttt{upscale\_method = bilinear}
            \item \texttt{scale\_by = 3.00}
        \end{itemize}
    \end{itemize}
    \item \textbf{ImageBlur}
    \begin{itemize}[label={}]
        \item \texttt{inputs:}
        \begin{itemize}[label={}]
            \item \texttt{image = Upscale Image By}
        \end{itemize}
        \item \texttt{parameters:}
        \begin{itemize}[label={}]
            \item \texttt{blur\_radius = 5}
            \item \texttt{sigma = 0.4}
        \end{itemize}
    \end{itemize}
    \item \textbf{Brightness \& Contrast}
    \begin{itemize}[label={}]
        \item \texttt{inputs:}
        \begin{itemize}[label={}]
            \item \texttt{image = ImageBlur}
        \end{itemize}
        \item \texttt{parameters:}
        \begin{itemize}[label={}]
            \item \texttt{contrast = 0.80}
            \item \texttt{brightness = 0.90}
        \end{itemize}
    \end{itemize}
\end{itemize}

\textbf{OUTPUT}

\begin{itemize}[label=--]
    \item \textbf{Save Image}
    \begin{itemize}[label={}]
        \item \texttt{inputs:}
        \begin{itemize}[label={}]
            \item \texttt{images = Brightness \& Contrast}
        \end{itemize}
        \item \texttt{parameters:}
        \begin{itemize}[label={}]
            \item \texttt{filename\_prefix = ComfyUI}
        \end{itemize}
    \end{itemize}
\end{itemize}

\clearpage

{\small
\bibliographystylesupp{bib/ieee_mine}
\bibliographysupp{bib/egbib_supp}
}
\end{document}